\pdfoutput=1
\PassOptionsToPackage{table,xcdraw,dvipsnames}{xcolor}

\documentclass[11pt]{article}

\usepackage[final]{coling}
\usepackage{times}
\usepackage{latexsym}
\usepackage{graphicx}
\usepackage{url} 
\usepackage{amsmath}
\usepackage{amsfonts} 
\usepackage{microtype} 
\usepackage{lipsum} 
\usepackage{stfloats}
\usepackage{multicol}
\usepackage{array}
\usepackage{multirow}
\usepackage[T1]{fontenc}

\usepackage{tcolorbox}
\usepackage{microtype}

\usepackage{inconsolata}

\usepackage{graphicx}

\usepackage{booktabs}

\title{Linguistic Minimal Pairs Elicit Linguistic Similarity \\ in Large Language Models}

\author{
\textbf{Xinyu Zhou\textsuperscript{1}}\thanks{Joint first authors. Xinyu Zhou is now affiliated with Université Paris Cité and Sorbonne Université.}\quad
\textbf{Delong Chen\textsuperscript{2}}$^*$\quad\\
\textbf{Samuel Cahyawijaya\textsuperscript{2,4}}\quad
\textbf{Xufeng Duan\textsuperscript{1}}\quad
\textbf{Zhenguang G. Cai\textsuperscript{1,3}}\\
\textsuperscript{1}Department of Linguistics and Modern Languages, CUHK\\
\textsuperscript{2}Department of Electronic and Computer Engineering, HKUST\\
\textsuperscript{3}Brain and Mind Institute, CUHK\\
\textsuperscript{4}Cohere\\
\texttt{xinyuzhou314@gmail.com},\texttt{ delong.chen@connect.ust.hk}
}

\begin{document}
\maketitle
\begin{abstract}
We introduce a novel analysis that leverages linguistic minimal pairs to probe the internal linguistic representations of Large Language Models (LLMs). By measuring the similarity between LLM activation differences across minimal pairs, we quantify the \textit{linguistic similarity} and gain insight into the linguistic knowledge captured by LLMs. Our large-scale experiments, spanning 100+ LLMs and 150k minimal pairs in three languages, reveal properties of linguistic similarity from four key aspects: consistency across LLMs, relation to theoretical categorizations, dependency to semantic context, and cross-lingual alignment of relevant phenomena.
Our findings suggest that 1)~linguistic similarity is significantly influenced by training data exposure, leading to higher cross-LLM agreement in higher-resource languages. 2)~Linguistic similarity strongly aligns with fine-grained theoretical linguistic categories but weakly with broader ones. 3)~Linguistic similarity shows a weak correlation with semantic similarity, showing its context-dependent nature. 4)~LLMs exhibit limited cross-lingual alignment in their understanding of relevant linguistic phenomena. This work demonstrates the potential of minimal pairs as a window into the neural representations of language in LLMs, shedding light on the relationship between LLMs and linguistic theory.
\end{abstract}

\begin{figure*}[!h]
    \centering
    \includegraphics[width=0.9\linewidth]{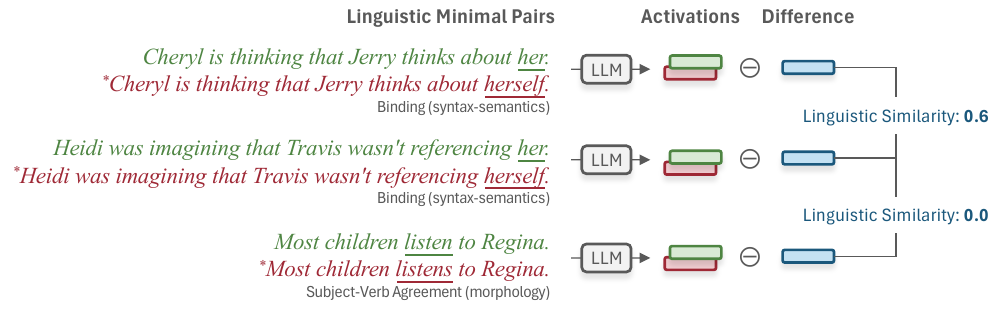}
    \caption{\textbf{The process of measuring linguistic similarity in an LLM}. We extract LLM activations for sentences in linguistic minimal pairs and compute their differences. Since the sentences differ solely in a specific linguistic phenomenon, the resulting difference only contains information about that phenomenon. We then measure the similarity between these activation differences, which we refer to as \textit{linguistic similarity}.}
    \label{fig:linguistic_similarity_calculation}
    \vspace{-10pt}
\end{figure*}

\section{Introduction}

The categorization of linguistic phenomena\footnote{Linguistic phenomena refer to observable patterns or features in language use. For example, subject-verb agreement is a linguistic phenomenon where verbs must agree with subjects in number and person. An example would be: \textit{``The dog barks''} (correct) instead of \textit{``*The dog bark''} (incorrect).} based on their relevance has been a long-standing endeavor, dating back to Aristotle~\cite{Aristotle350BC}. This has led to the widely accepted theoretical linguistic consensus of a hierarchical categorization of language structure encompassing syntax, semantics, morphology, etc., which provides a structured way to understand the intricate nature of language, and allows linguists to investigate the interrelationships and commonalities among these linguistic domains~\cite{Comorovski2013,Li2004}.

Alongside the theoretical discussions of linguistic phenomena, a growing body of research on \textit{quantitative measurement of similarities} based on statistical modeling on large-scale corpora has been observed in computational linguistics. Examples include lexical similarity~\cite{Holman2011}, syntactic similarity~\cite{Boghrati2018, Schoot2016}, semantic similarity~\cite{Pennington2014,Reimers2019}, among others. These examples showcase the possibilities of understanding the nature of language through purely statistical methodologies. However, there has been limited research on quantitatively measuring the relationships between different linguistic phenomena. Given that language is a complex system composed of numerous interrelated linguistic phenomena, addressing this gap could lead to a more comprehensive understanding of language structure and its underlying mechanisms.

In this work, we aim to uncover and analyze the internal linguistic knowledge of Large Language Models (LLMs) when presented with a wide range of linguistic phenomena. LLMs are large-scale unsupervised language learners without any prior linguistic knowledge, and have demonstrated human-level language capability, as evidenced by their leading performance on language understanding benchmarks and impressive language generation fluency~\cite{Zhao2023,Bang2023}. More specifically, we are interested in how LLMs represent different linguistic phenomena, and whether linguistically similar phenomena are represented similarly in LLMs.

To elicit such representations, we examine the activations in LLMs in response to linguistic minimal pairs~\cite{Warstadt2020}. As shown in Fig.~\ref{fig:linguistic_similarity_calculation}, these pairs consist of sentences that differ only in a word/phrase, with one being grammatical and the other ungrammatical. Since minimal pairs differ \textit{only} in one particular linguistic phenomenon, information about other aspects (such as topic and semantic meaning) will be canceled out through subtraction. We interpret the remaining differences as the LLMs' internal representation of a specific linguistic phenomenon. By calculating the similarity between multiple such representations, we derive a measure of \textit{linguistic similarity} between linguistic minimal pairs.

We then conduct an extensive analysis of linguistic similarity in LLMs. Our experiment encompasses 100+ LLMs of varying scales and pretraining corpora, utilizing 150,000 linguistic minimal pairs across 3 different languages. We report our observations correspond to the following key questions:

\textbf{1) How consistent is linguistic similarity across different LLMs?} 
LLMs have the highest alignment of linguistic similarity in English, which is the most widely used language for LLM pertaining, while the alignments are comparatively weaker in Chinese and Russian. We further visualized the relationships among these LLMs with UMAP~\cite{McInnes2018}. On Chinese samples, we observed a distinct clustering pattern: bilingual and multilingual LLMs formed one cluster, while English-only models formed another. The above results suggest that the language distribution in the training data influences the linguistic similarity in LLMs.

\textbf{2) Does linguistic similarity align with theoretical linguistic categorizations?} We compared linguistic similarity across three levels of theoretical linguistic categorizations. Our analysis revealed that fine-grained classifications exhibit significantly higher intra-class similarities compared to inter-class similarities. However, this disparity diminishes considerably at higher categorization levels. 
Meanwhile, we can also observe some highly correlated phenomena pairs that are not classified to the same theoretical categorization.

\textbf{3) To what extent does linguistic similarity correlate with semantic similarity?} We showed a weak correlation between semantic similarity and linguistic similarity, despite many existing samples with low linguistic similarity and high semantic similarity, and conversely, high in linguistic and low in semantic. The weak correlation indicates that linguistic similarity in LLMs has a \textit{context-dependent} nature.

\textbf{4) Whether relevant phenomena in different languages enjoy higher linguistic similarities?} We compare the linguistic similarity of the shared three linguistic phenomena in English and Chinese. 
Our UMAP visualization revealed that while English phenomena are clustered within a shared region, they are ``attracted'' by their relevant phenomena in Chinese.


We hope this paper sparks new exploration into LLMs' internal linguistic representations, uncovering deeper insights into their inner workings and potentially informing linguistic theory. 
To facilitate future research, the activation differences of the 100+ LLMs, pre-computed sample-level linguistic similarities, and all the codes are made publicly available at \url{https://github.com/ChenDelong1999/Linguistic-Similarity}.

\section{Related Work}

\textbf{Linguistic Minimal Pairs}.
Acceptability judgments~\cite{Chomsky1957} have long served as a proxy for grammaticalness in generative syntax, relying on native speakers' intuitions to evaluate whether sequences generated by a grammar are perceived as acceptable. In the past decade, grammatical acceptability judgment tasks~\cite{Linzen2016, Futrell2018, Wilcox2019, Warstadt2019, Gauthier2020} have been commonly used to evaluate language models by comparing output probabilities, providing a direct linguistic measure of sentence acceptability. Among these benchmarks, BLiMP~\cite{Warstadt2020} introduced a large-scale linguistic minimal pair benchmark in English. This work was followed by many studies in other languages, namely CLiMP~\cite{Xiang2021} and SLING~\cite{Song2022} for Chinese, JBLiMP~\cite{Someya2023} for Japanese, BHASA~\cite{Leong2023} for Indonesian, RuBLiMP~\cite{Taktasheva2024} for Russian and BLiMP-NL~\cite{Suijkerbuijk2024} for Dutch.

\textbf{Linguistic and Language Representation in LLMs}.
A growing body of research is investigating the linguistic mechanisms within LLMs through probing and interventional strategies~\cite{Arora2024, He2024, Duan2024, Weber2024}. These studies mostly focus on specific linguistic phenomena, such as subject-verb agreement~\cite{Giulianelli2018}, plurality~\cite{Hanna2022}, long-distance agreement~\cite{Li2023}, negative polarity items~\cite{DeCarlo2023}, and adjective order~\cite{Jumelet2024}, see~\citet{Milliere2024} for a comprehensive review. Furthermore, recent studies have also identified linguistic regions~\cite{Zhang2024} and language-specific neurons~\cite{Tang2024, Kojima2024} that contribute to multilingual capabilities, and further suggests that LLMs exhibit layer-wise specialization, with intermediate layers processing information in a common "language" concept space and final layers generating responses in the specific language~\cite{Wendler2024,Zhong2024}.

\section{Measuring Linguistic Similarity in Large Language Models}

\subsection{Definition}

Let $x^+$ denote a grammatically correct natural language sentence, and $x^-$ represent an ungrammatical sentence derived from $x^+$ with a minimal modification affecting a specific linguistic phenomenon. The pair $\langle x^+, x^- \rangle$ is referred to as a \textit{linguistic minimal pair}. Let $f_\text{LLM}$ denote an LLM that takes sentences as input and generates corresponding hidden activations, \textit{i.e.,} $z^+ = f_\text{LLM}(x^+)$ and $z^- = f_\text{LLM}(x^-)$, where $z^+, z^- \in \mathbb{R}^n$ and $n$ is the dimensionality of the hidden representations.

We compute the difference between the hidden activations: $\Delta z = z^+ - z^-$. While $z^+$ and $z^-$ individually encode rich information about the input sentences, including both semantic and linguistic properties, their difference $\Delta z$ primarily captures the representation of the specific linguistic phenomenon that distinguishes the minimal pair, as other aspects of the sentences are guaranteed to be identical and thus will be canceled out. Formally, given two linguistic minimal pairs $\langle x^+_1, x^-_1 \rangle$ and $\langle x^+_2, x^-_2 \rangle$, we define their linguistic similarity as $\texttt{sim}(\Delta z_1, \Delta z_2)$, where $\texttt{sim}(\cdot)$ is a similarity metric and we used cosine similarity in this work. 

\subsection{Implementation}

\textbf{Data.} 
We utilized linguistic minimal pairs from three existing datasets: BLiMP~\cite{Warstadt2020}, SLING~\cite{Song2022}, and RuBLiMP~\cite{Taktasheva2024}, which consist of 67, 38, and 45 linguistic phenomena in English, Chinese, and Russian, respectively. Each linguistic phenomenon is associated with 1,000 corresponding linguistic minimal pairs, yielding a total of 150,000 pairs.

\textbf{LLMs.} 
Each sentence was input into various LLMs without any prompts, and hidden states were extracted from five evenly sampled layers. We specifically extracted the activations of the last-but-two token, as previous tokens remain visible, while the final tokens correspond to the \texttt{<end-of-sentence>} token. A comprehensive range of 104 LLMs from Huggingface was employed; the complete list is provided in Appendix~\ref{appendix:full_list_llms}. 

\textbf{Computation.} 
All inferences on the LLMs were conducted using half-precision (\texttt{float-16}). Given the extensive volume of linguistic minimal pairs and the number of LLMs, we optimized storage by saving both the activation differences and the computed pairwise linguistic similarity matrix in \texttt{int8} precision. Under this configuration, the activation differences for \texttt{Llama-2-7B} required 1.3 GB of storage (a tensor of 67,000 samples $\times$ 5 layers $\times$ 4096 neurons), while the similarity matrix of 67,000 $\times$ 67,000 necessitated 4.2 GB of storage.

\section{Result and Discussion}

\subsection{Consistency of Linguistic Similarity Across 104 LLMs}
\label{sec:Align_LLMs}

\begin{figure*}
    \centering
    \includegraphics[width=0.8\linewidth]{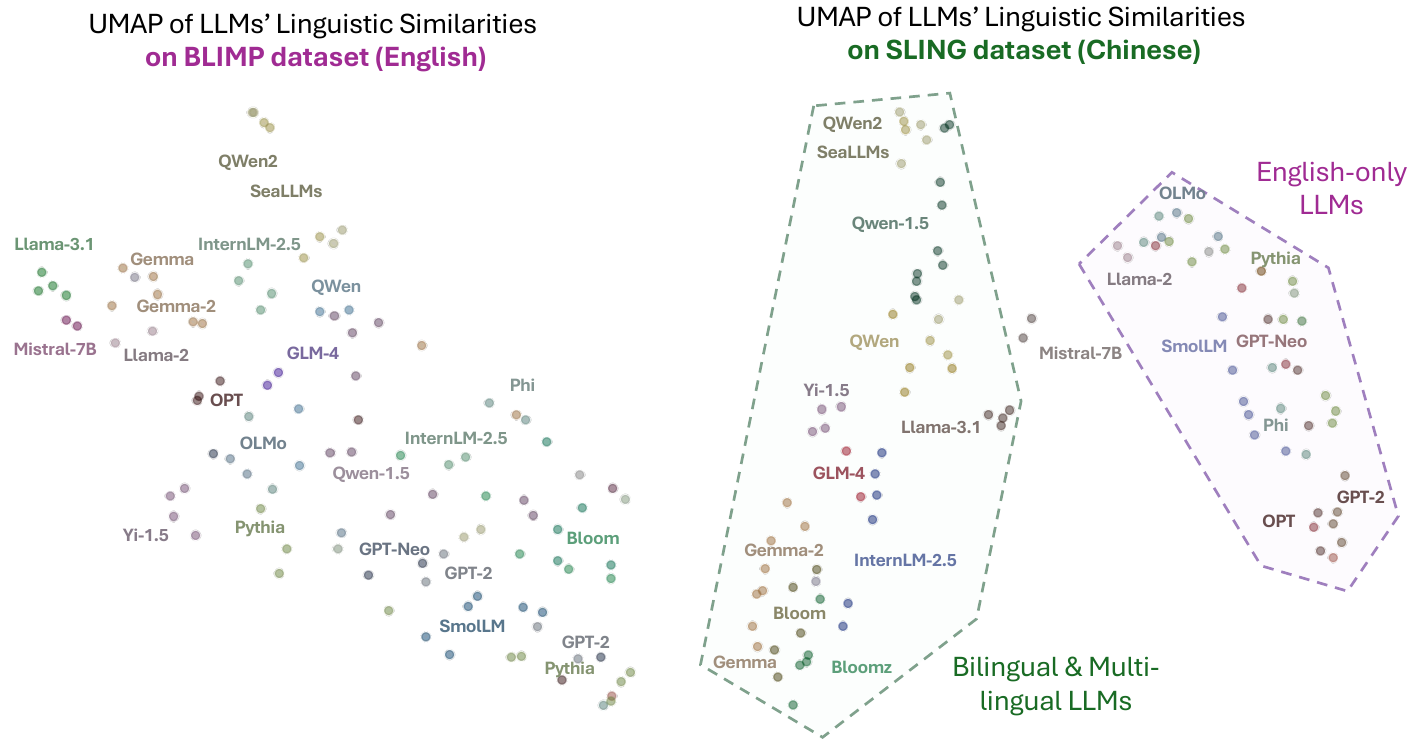}
    \caption{\textbf{The relationship between linguistic similarity across LLMs}. In English, LLMs form a single cluster, while in Chinese, two distinct clusters emerge: one for bilingual and multilingual LLMs, and another for English-only models. Detailed visualizations can be found in Appendix~\ref{appendix:llm_alignment}.
    }
    \label{fig:LLM_alignments}
    \vspace{-10pt}
\end{figure*}

To quantitatively assess the consistency of linguistic similarity across different LLMs, we adopted the \textit{mutual k-nearest neighbors} metric as proposed in~\cite{Huh2024}. Specifically, for a given linguistic minimal pair, we retrieved the top-k closest neighbors based on linguistic similarity from two distinct LLMs and calculated the percentage of overlapping samples among the retrieved sets as the alignment score (see \citealp{Huh2024} for further details).

\begin{figure}[!t]
    \centering
    \includegraphics[width=1\linewidth]{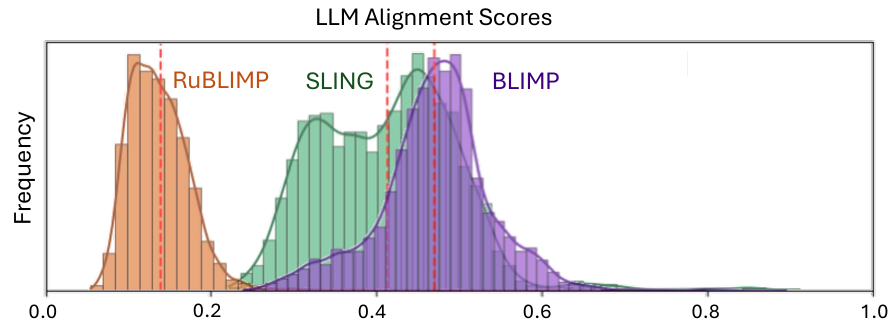}
    \caption{Distribution of LLM alignment scores, with red dotted lines marks the average scores
of 0.471 (BLiMP), 0.414 (SLING), and 0.139 (RuBLiMP).}
    \label{fig:LLM_alignments_distribution}
    \vspace{-10pt}
\end{figure}

We computed pairwise alignment scores for all 104 LLMs across three datasets. For computational efficiency, we randomly sampled 10\% of the total samples from each dataset (\textit{i.e.,} 6,700 for BLiMP, 3,800 for SLING, and 4,500 for RuBLiMP) and set $k$ to 1\% of the sample pool size for retrieval (\textit{i.e.,} 67, 38, and 45). The distribution of alignment scores is presented in Fig.~\ref{fig:LLM_alignments_distribution}. Notably, the BLiMP dataset (English) exhibited the highest alignment, with an average score of 0.471 (\textit{i.e.,} 47.1\% of the top-1\% similar minimal pairs are shared across LLMs on average), whereas the SLING (Chinese) and RuBLiMP (Russian)  demonstrated average alignment scores of 0.414 and 0.139, respectively. 

We also observed that the distribution of alignment scores for English and Russian datasets is unimodal, while the Chinese dataset exhibits a bimodal distribution. To further investigate this phenomenon, we employed UMAP~\cite{McInnes2018} to embed the LLMs to 2D plane based on a distance metric of \texttt{distance} $= -\log(\texttt{alignment score})$, following \citet{Huh2024}. As illustrated in Fig.~\ref{fig:LLM_alignments}, each dot represents an LLM, with closer dots indicating similar linguistic similarities. LLMs from the same family (\textit{e.g.,} models from the same creator with different sizes or base/chat versions) are clustered together, indicated by the same color. Interestingly, in Chinese, we observed two distinct clusters: the left cluster (marked with a green dotted line) predominantly consists of bilingual (English-Chinese) and multilingual models, while the right cluster marked with purple is primarily composed of English-only trained LLMs. These results suggest a correlation between alignment scores and LLMs' pertaining language.

\begin{tcolorbox}[
  title=Takeaway:,
  fonttitle=\bfseries,
  colbacktitle=gray!20,
  colframe=white,
  colback=gray!4,
  coltitle=black
]
\textit{LLMs' linguistic similarity is dependent on their training data exposure. They show stronger agreements in higher-resource languages like English.}
\end{tcolorbox}

\begin{figure*}
    \centering
    \includegraphics[width=1\linewidth]{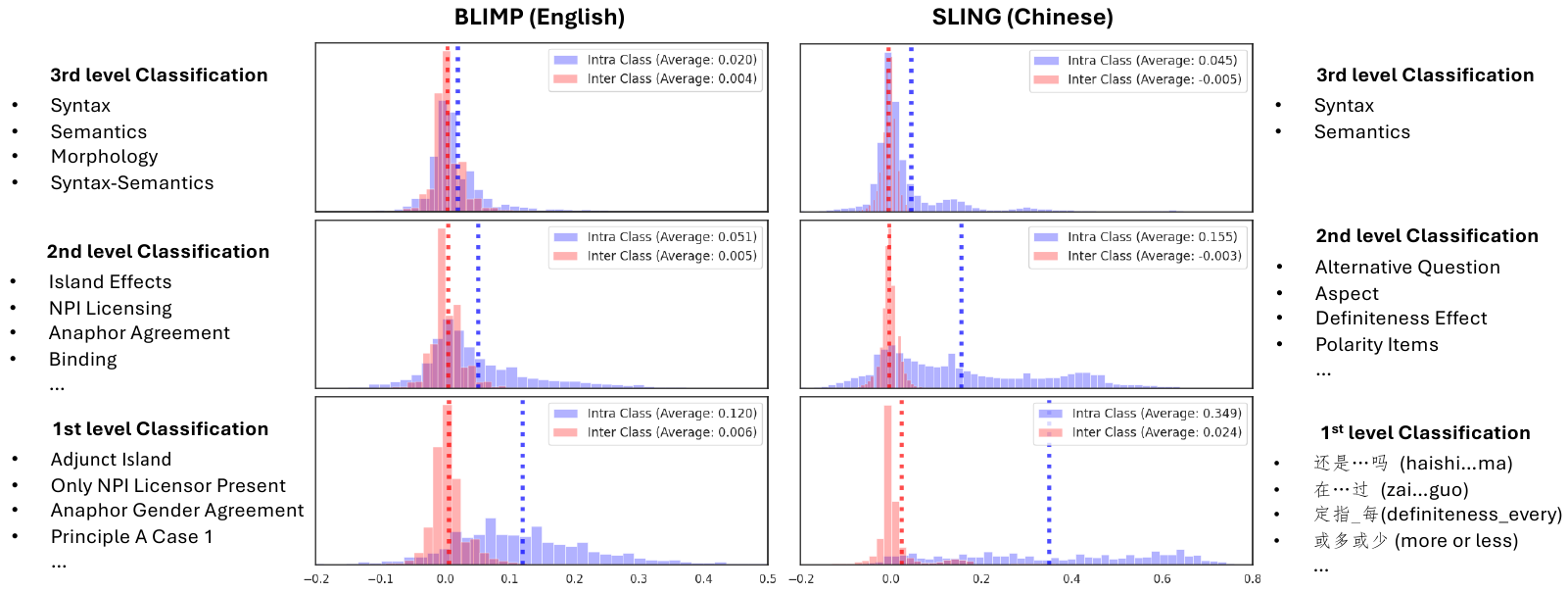}
     \caption{\textbf{Intra-class and inter-class linguistic similarities at different levels of linguistic classification}. At the most fine-grained level (1st level), intra-class similarities are significantly higher than inter-class similarities, indicating a strong alignment with detailed theoretical linguistic categorizations. As we move to broader categories (2nd and 3rd levels), the gap between inner and inter-class similarities narrows notably.}
    \label{fig:linguistic_similarity_distributions}
\end{figure*}

\subsection{Alignment between Linguistic Similarity and Theoretical Categorizations}

To investigate the alignment between linguistic similarity and theoretical linguistic categorizations, we compared intra-class and inter-class similarities at different levels of linguistic classification for both English (BLiMP) and Chinese (SLING) datasets. BLiMP and SLING provide hierarchical linguistic classifications, with the 1st level being the most fine-grained ``phenomena'', the 2nd level capturing broader categories of ``terms'', and the 3rd level representing the most general ``fields'' (examples are shown in leftmost and rightmost columns in Fig.~\ref{fig:linguistic_similarity_distributions}).

The analysis in this section is based on averaged linguistic similarities across the 104 LLMs. We focused on BLiMP and SLING datasets due to their high linguistic similarity alignment across LLMs, as demonstrated in the previous Section~\ref{sec:Align_LLMs}. We excluded the RuBLiMP (Russian) dataset with weaker consensus across LLMs. 

As shown in Fig.~\ref{fig:linguistic_similarity_distributions}, our analysis revealed that at the lowest level, intra-class similarities were significantly higher than inter-class similarities for both BLiMP and SLING datasets. This suggests that linguistic similarity effectively captures the nuanced distinctions within these detailed categories. This clear separation indicates a strong alignment between linguistic similarity and these fine-grained theoretical linguistic categorizations. However, as we moved to higher levels of classification, the gap between intra-class and inter-class similarities diminished considerably. For both datasets, the disparity between inner and inter-class similarities approximately halved with each ascent in the categorization hierarchy. Higher-level linguistic categories exhibit greater interconnectedness and mutual influence than previously recognized, which may explain the diminishing differentiation in linguistic similarity at broader classification levels.

\begin{figure*}[t!]
    \centering
    \includegraphics[width=1\linewidth]{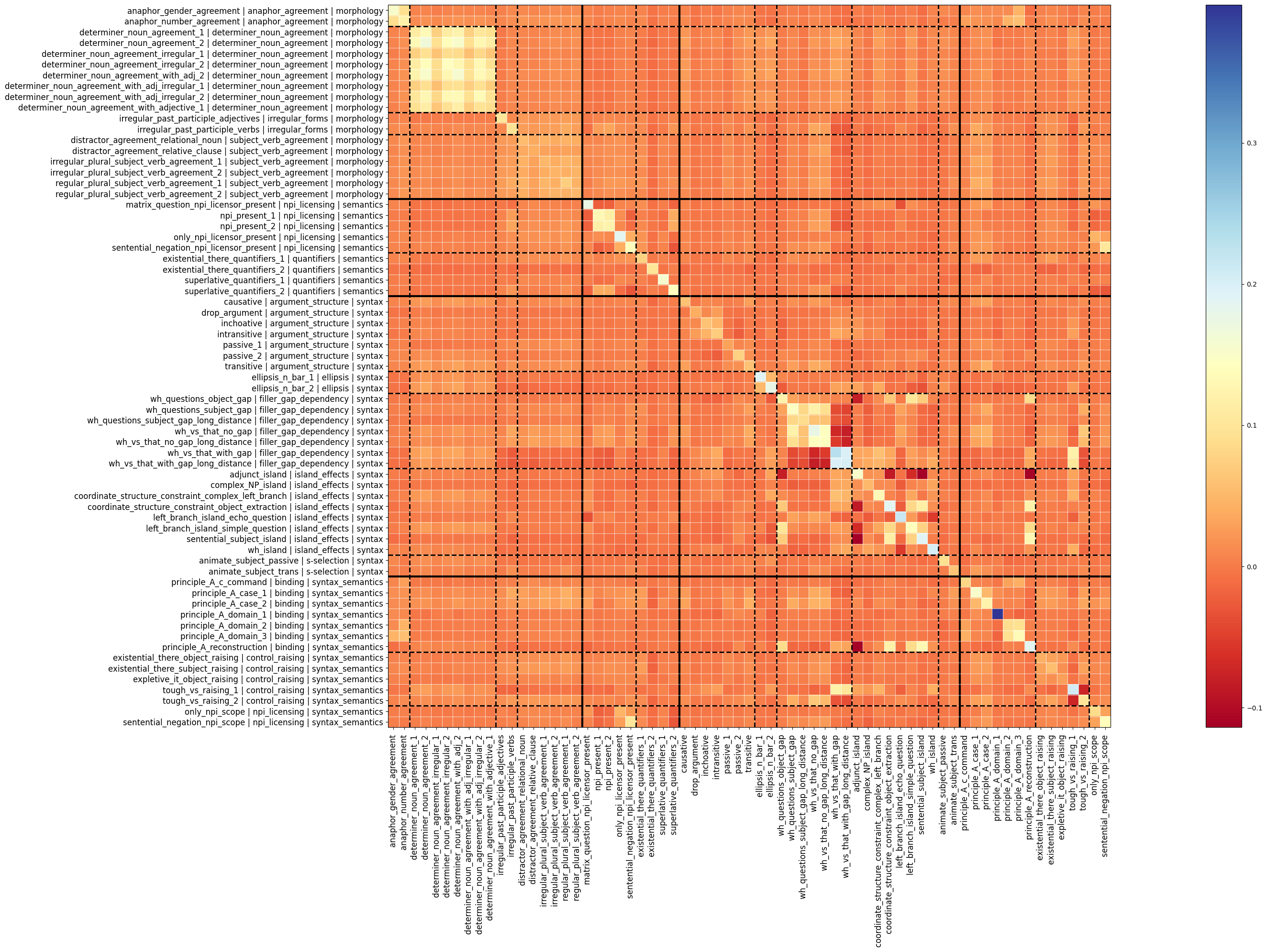}
     \caption{\textbf{Phenomena-level linguistic similarity matrix of BLiMP}. Each grid corresponds to the average similarity between two linguistic phenomena. The categorization in the 2nd-level (linguistic terms) and the 3rd-level are respectively separated by dashed and bold black lines. On the left, we provide label of the 1st to 3rd levels of linguistic classifications, separated by ``|''. Visualizations of SLING and RuBLiMP can be found in Appendix~\ref{appendix:Phenomena Similarities}.}
    \label{fig:phenomena_correlations_BLiMP}
\end{figure*}

We further visualize the phenomenon-level linguistic similarity matrix in Fig.~\ref{fig:phenomena_correlations_BLiMP} (visualizations for other datasets can be found in Appendix~\ref{appendix:Phenomena Similarities}). The observations here confirm our above conclusions from Fig.~\ref{fig:linguistic_similarity_distributions}. The clearly distinguishable diagonal entries represent the lowest linguistic phenomenon-level similarity, which enjoys significant inter/intra separation. For higher 2nd-level classification, as noted by dashed black lines, there exist both homogeneous (\textit{e.g.,} anaphor agreement, determiner none agreement) and heterogeneous (\textit{e.g.,} irregular forms and many others) ones. We cannot observe any clear clustering effects for the 3rd-level classification as separated by bold black lines, which also aligns with our findings from  Fig.~\ref{fig:linguistic_similarity_distributions}. 

Interestingly, we found that there are some pairs of phenomena that do not belong to the exact same theoretical categorization, but also enjoy considerably high similarities. For example, \textit{``sentential negation NPI licensor present''} in the \textit{``NPI licensing''} term of \textit{``semantics''} field has very high similarity with \textit{``sentential negation NPI scope''} in the \textit{``NPI licensing''} term of \textit{``syntax-semantics''} field; and \textit{``principle A domain 3''} in the \textit{``binding''} term of \textit{``syntax-semantics''} filed has considerably high similarities to the phenomena in \textit{``anaphor agreement''} term of the \textit{``morphology''} field. These observations underscore the potential of linguistic similarity as a valuable tool for refining our understanding of language structure and organization.

\begin{tcolorbox}[
  title=Takeaway:,
  fonttitle=\bfseries,
  colbacktitle=gray!20,
  colframe=white,
  colback=gray!4,
  coltitle=black
]
\textit{Linguistic similarity in LLMs aligns well with fine-grained theoretical categorizations, but weakens at higher levels.
}
\end{tcolorbox}

\subsection{The Relationship Between Linguistic Similarity and Semantic Similarity}

\begin{figure*}
    \centering
    \includegraphics[width=1\linewidth]{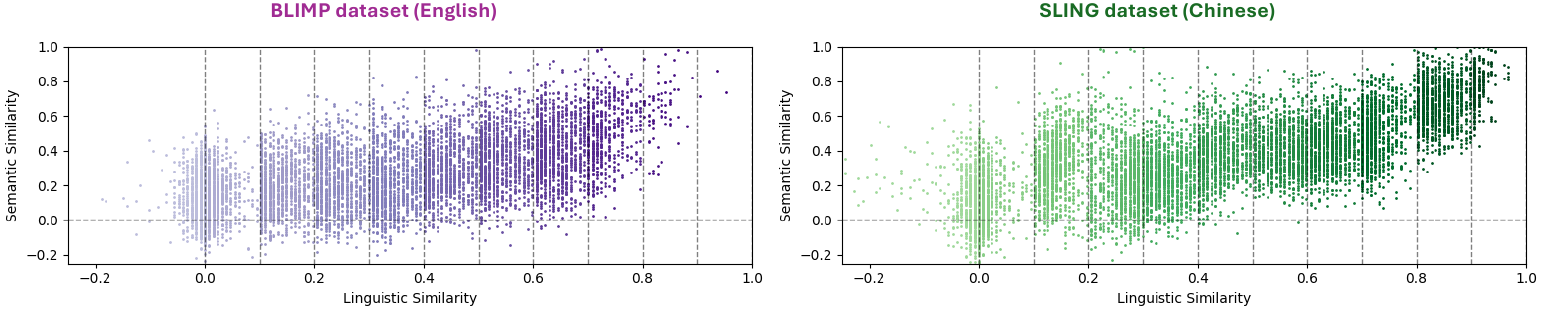}
     \caption{\textbf{Joint distribution of linguistic similarity and semantic similarity}. Each dot in the plots represents a pair of linguistic minimal pairs. We observed a weak correlation between the two similarity measurements, showing that linguistic similarity has a context-dependent nature.}
    \label{fig:linguistic_semantic}
\end{figure*}

\begin{table*}
    \centering
    \includegraphics[width=1\linewidth]{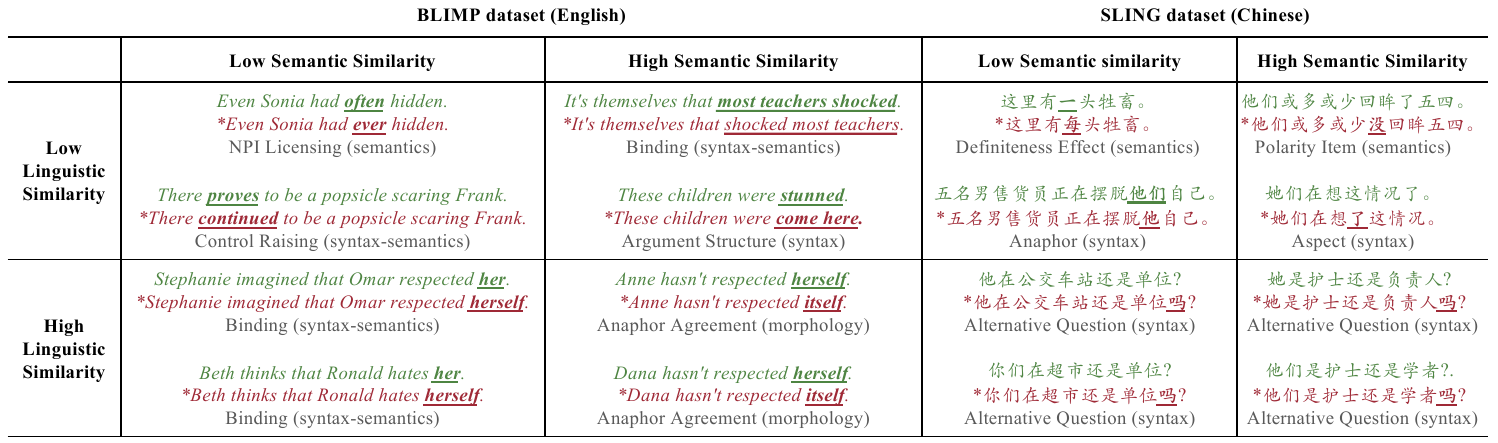}
     \caption{\textbf{Examples of pairs of minimal pairs that have different levels of linguistic and semantic similarity}. ``High'' means similarity is larger than 0.6 while ``low'' means smaller than 0.3. Although linguistic similarity exhibits a weak correlation to semantic similarity and is context-dependent, there are also many samples with low linguistic similarity and high semantic similarity, and conversely, high in linguistic and low in semantic, in both English and Chinese.}
    \label{fig:linguistic_semantic_table}
\end{table*}

To further investigate the nature of linguistic similarity captured by our method, we compare it with the semantic similarity between minimal pairs. We employed a multilingual Sentence Transformer~\cite{Reimers2019} model\footnote{\url{https://huggingface.co/sentence-transformers/paraphrase-multilingual-MiniLM-L12-v2}} to generate sentence embeddings for the correct sentence in the minimal pairs. Cosine similarity between these embeddings served as our measure of semantic similarity following default practice. We sampled 1k pairs of minimal pairs that have linguistic similarity within each range of (0.9, 1.0), (0.8, 0.9), ..., (0, 0.1), ($-\infty$, 0), and plotted their semantic similarity against their linguistic similarity. 

Fig.~\ref{fig:linguistic_semantic} provides the results for the BLiMP and SLING. We can observe a weak correlation between linguistic and semantic similarities for both datasets, suggesting that the linguistic similarity in LLMs is \textit{context dependent} to some extent. However, as can be seen in Table~\ref{fig:linguistic_semantic_table}, linguistic and semantic similarities can vary independently. In English, two minimal pairs from the Binding (syntax-semantics) phenomenon show high linguistic similarity (>0.6) but low semantic similarity (<0.3), indicating that while these pairs involve the same linguistic structures, the semantic impact of the changes differs substantially. Conversely, we found cases of low linguistic similarity (<0.3) but high semantic similarity (>0.6) between minimal pairs from Binding (syntax-semantics) and Argument Structure (syntax). We also observed cases where both linguistic and semantic similarities are consistently high or low, which often emerges when minimal pairs share similar vocabulary or address completely unrelated linguistic features. The result of SLING reveals similar patterns to those observed in BLiMP.

\begin{tcolorbox}[
  title=Takeaway:,
  fonttitle=\bfseries,
  colbacktitle=gray!20,
  colframe=white,
  colback=gray!4,
  coltitle=black
]
\textit{LLMs' linguistic similarity shows a weak correlation with semantic similarity, indicating that it is context-dependent.}
\end{tcolorbox}

\subsection{Linguistic Similarity Across Different Languages}
\label{sec:across-languages}

We are interested in how LLMs represent similar linguistic phenomena across different languages--for relevant phenomena in different languages, whether they exhibit higher linguistic similarity? We conducted a multi-lingual analysis focusing on three key linguistic terms: anaphor agreement, polarity items, and filler-gap dependency, on both BLiMP (English) and SLING (Chinese) datasets. Linguistic phenomena within each of these terms are considered relevant, and a total of 39 phenomena are involved, leading to a total of 39k minimal pairs. 

\begin{figure*}
    \centering
    \includegraphics[width=1\linewidth]{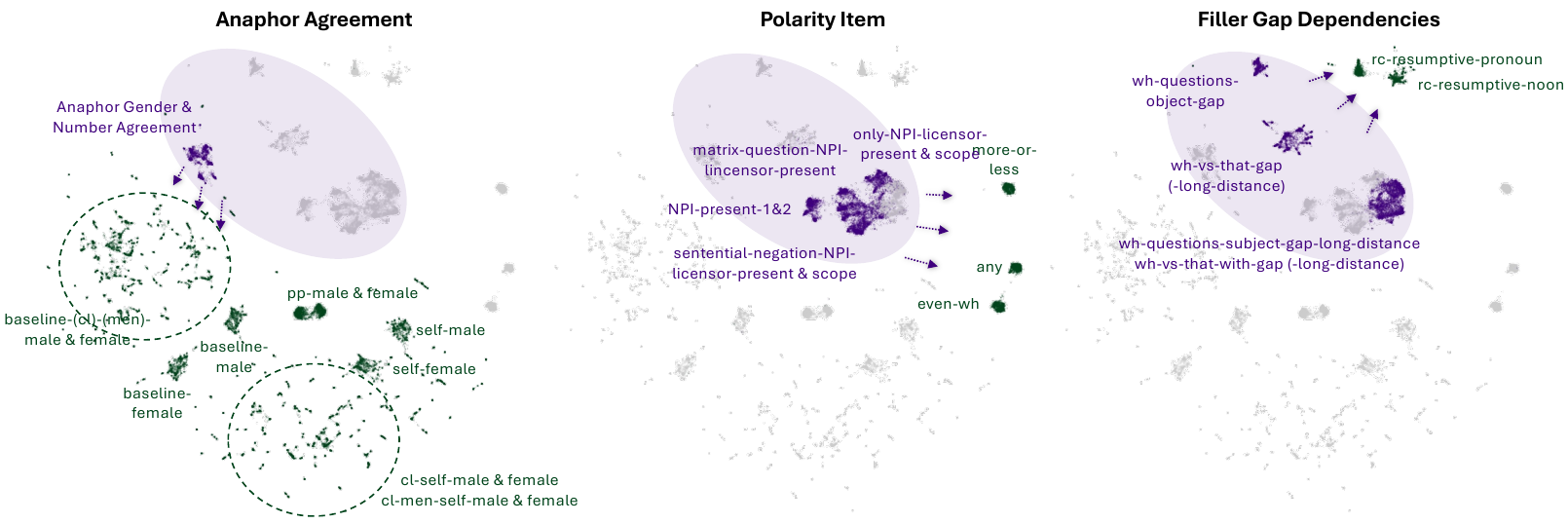}
     \caption{\textbf{UMAP visualization of minimal pairs in same categories (terms) but different languages}. Linguistic phenomena are dominantly grouped by their language in multilingual LLMs: English samples are all clustered within purple-shaded areas. While relevant linguistic phenomena in different languages are not fully overlapped, LLM does capture some relationships. As indicated by purple arrows, English samples seem to be ``attracted'' by the corresponding Chinese samples.}
    \label{fig:cross_lingual_umap}
\end{figure*}

To explore the relationship between those phenomena, we used a representative state-of-the-art multilingual LLM \texttt{Llama-3.1}\footnote{\url{https://huggingface.co/meta-llama/Meta-Llama-3.1-8B}}, and employed UMAP dimensionality reduction based on pair-wise linguistic similarity. Figure~\ref{fig:cross_lingual_umap} visualizes the results. We observed a clear \textit{language-specific clustering} pattern, as evident by the purple-shaded area where all English samples are clustered. It means that relevant linguistic phenomena in different languages are considered to be \textit{different} by multilingual LLMs. 

Interestingly, we also observe that relevant phenomena across different languages often exhibit closer proximity than unrelated phenomena within the same language. This cross-lingual correlation is illustrated by the purple arrows in Figure~\ref{fig:cross_lingual_umap}, which \textit{``attracts''} English samples to Chinese samples. Quantitatively, we calculated the average pair-wise linguistic similarity of the three terms in BLiMP and SLING. As Table~\ref{tab:cross-lingual}, for each term in English (\textit{i.e.,} each column), the corresponding Chinese term always enjoys the highest similarity. This ``attraction'' pattern suggests that LLMs may have captured some relationships of these linguistic structures that transcend individual language boundaries.


\begin{table}[]
\caption{Average linguistic similarity of phenomena in English and Chinese.}
\label{tab:cross-lingual}
\resizebox{\linewidth}{!}{%
\begin{tabular}{cc|ccc}
\toprule
\multicolumn{1}{l}{} &
  \multicolumn{1}{l|}{} &
  \multicolumn{3}{c}{{\color[HTML]{674EA7} \textbf{BLiMP dataset (English)}}} \\
\multicolumn{1}{l}{} &
  \multicolumn{1}{l|}{} &
  \multicolumn{1}{c|}{\cellcolor[HTML]{EEECF5}{\color[HTML]{000000} \begin{tabular}[c]{@{}c@{}}Anaphor\\ Agreement\end{tabular}}} &
  \multicolumn{1}{c|}{\cellcolor[HTML]{EEECF5}{\color[HTML]{000000} \begin{tabular}[c]{@{}c@{}}Polarity\\ Item\end{tabular}}} &
  \cellcolor[HTML]{EEECF5}{\color[HTML]{000000} \begin{tabular}[c]{@{}c@{}}Filler Gap\\ Dependency\end{tabular}} \\ \hline
{\color[HTML]{274E13} \textbf{SLING}} &
  \cellcolor[HTML]{EDF5EB}{\color[HTML]{000000} Anaphor Agreement} &
  \multicolumn{1}{c|}{\textbf{.04853}} &
  \multicolumn{1}{c|}{.00643} &
  -.01850 \\
{\color[HTML]{274E13} \textbf{Dataset}} &
  \cellcolor[HTML]{EDF5EB}{\color[HTML]{000000} Polarity Item} &
  \multicolumn{1}{c|}{.01883} &
  \multicolumn{1}{c|}{\textbf{.01783}} &
  .00714 \\
{\color[HTML]{274E13} \textbf{(Chinese)}} &
  \cellcolor[HTML]{EDF5EB}{\color[HTML]{000000} Filler Gap Dependency} &
  \multicolumn{1}{c|}{.02735} &
  \multicolumn{1}{c|}{.01780} &
  \textbf{.01426} \\ \bottomrule
\end{tabular}%

}
\end{table}

\begin{tcolorbox}[
  title=Takeaway:,
  fonttitle=\bfseries,
  colbacktitle=gray!20,
  colframe=white,
  colback=gray!4,
  coltitle=black
]
\textit{Linguistic phenomena are grouped by language in multilingual LLMs. Relevant phenomena in different languages are not fully overlapped, but their relevance is indeed being captured.}
\end{tcolorbox}

\section{Conclusion}

Our comprehensive analysis of linguistic similarity in LLMs, spanning over 100 models and 150k minimal pairs across three languages, reveals several key insights. We found that linguistic similarity consistency across LLMs is strongly influenced by pertaining data, with high-resource languages showing greater alignment. LLMs' internal representations align well with fine-grained linguistic categorizations, but this alignment weakens at broader levels. The weak correlation between linguistic and semantic similarities suggests that LLMs' representation of linguistic phenomena is context-dependent. Cross-lingually, while LLMs tend to group phenomena by language, they do capture some relationships between relevant phenomena across languages. 

These findings contribute to our understanding of LLMs' internal language processing, potentially bridging the gap between neural language models and linguistic theory. As LLMs continue to advance, this work provides a foundation for future research into their linguistic representations, informing model development and offering insights into both artificial and human language processing.




\section{Limitations and Future Works}

Our study's findings are inherently dependent on the quality and scope of existing linguistic minimal pair datasets, which may have the possibility of violating the assumption of having \textit{only} difference of individual phenomena. Additionally, our analysis is limited to three languages, which cannot fully represent global linguistic diversity. Despite examining 150 linguistic phenomena, this still captures only a fraction of language's complexity. Future directions include developing more high-quality minimal pairs, broadening language coverage to include more diverse linguistic families, and increasing the range of phenomena studied. 

\bibliography{reference}

\newpage
\appendix


\section{The List of 104 LLMs}
\label{appendix:full_list_llms}

The following is a list of all LLMs that are adopted in our analysis, grouped by model families:

\bigskip

{\small
\textbf{Llama-2}
\begin{itemize}
    \item \href{https://huggingface.co/NousResearch/Llama-2-7b-chat-hf}{\texttt{\textbf{NousResearch/Llama-2-7b-chat-hf}}}: Total 33 layers, 4096 neurons per layer. Sampled layers: 5, 11, 16, 22, 27.
    \item \href{https://huggingface.co/NousResearch/Llama-2-13b-chat-hf}{\texttt{\textbf{NousResearch/Llama-2-13b-chat-hf}}}: Total 41 layers, 5120 neurons per layer. Sampled layers: 6, 13, 20, 27, 34.
\end{itemize}

\textbf{Llama-3 \& Llama-3.1}
\begin{itemize}
    \item \href{https://huggingface.co/meta-llama/Meta-Llama-3-8B}{\texttt{\textbf{meta-llama/Meta-Llama-3-8B}}}: Total 33 layers, 4096 neurons per layer. Sampled layers: 5, 11, 16, 22, 27.
    \item \href{https://huggingface.co/meta-llama/Meta-Llama-3-8B-Instruct}{\texttt{\textbf{meta-llama/Meta-Llama-3-8B-Instruct}}}: Total 33 layers, 4096 neurons per layer. Sampled layers: 5, 11, 16, 22, 27.
    \item \href{https://huggingface.co/meta-llama/Meta-Llama-3.1-8B}{\texttt{\textbf{meta-llama/Meta-Llama-3.1-8B}}}: Total 33 layers, 4096 neurons per layer. Sampled layers: 5, 11, 16, 22, 27.
    \item \href{https://huggingface.co/meta-llama/Meta-Llama-3.1-8B-Instruct}{\texttt{\textbf{meta-llama/Meta-Llama-3.1-8B-Instruct}}}: Total 33 layers, 4096 neurons per layer. Sampled layers: 5, 11, 16, 22, 27.
\end{itemize}

\textbf{Mistral-7B-v0.3}
\begin{itemize}
    \item \href{https://huggingface.co/mistralai/Mistral-7B-v0.3}{\texttt{\textbf{mistralai/Mistral-7B-v0.3}}}: Total 33 layers, 4096 neurons per layer. Sampled layers: 5, 11, 16, 22, 27.
    \item \href{https://huggingface.co/mistralai/Mistral-7B-Instruct-v0.3}{\texttt{\textbf{mistralai/Mistral-7B-Instruct-v0.3}}}: Total 33 layers, 4096 neurons per layer. Sampled layers: 5, 11, 16, 22, 27.
\end{itemize}

\textbf{OLMo}
\begin{itemize}
    \item \href{https://huggingface.co/allenai/OLMo-1B-hf}{\texttt{\textbf{allenai/OLMo-1B-hf}}}: Total 17 layers, 2048 neurons per layer. Sampled layers: 2, 5, 8, 11, 14.
    \item \href{https://huggingface.co/allenai/OLMo-7B-hf}{\texttt{\textbf{allenai/OLMo-7B-hf}}}: Total 33 layers, 4096 neurons per layer. Sampled layers: 5, 11, 16, 22, 27.
    \item \href{https://huggingface.co/allenai/OLMo-7B-Instruct-hf}{\texttt{\textbf{allenai/OLMo-7B-Instruct-hf}}}: Total 33 layers, 4096 neurons per layer. Sampled layers: 5, 11, 16, 22, 27.
\end{itemize}

\textbf{Qwen, Qwen-1.5, \& Qwen-2}
\begin{itemize}
    \item \href{https://huggingface.co/Qwen/Qwen-7B}{\texttt{\textbf{Qwen/Qwen-7B}}}: Total 33 layers, 4096 neurons per layer. Sampled layers: 5, 11, 16, 22, 27.
    \item \href{https://huggingface.co/Qwen/Qwen-7B-Chat}{\texttt{\textbf{Qwen/Qwen-7B-Chat}}}: Total 33 layers, 4096 neurons per layer. Sampled layers: 5, 11, 16, 22, 27.
    \item \href{https://huggingface.co/Qwen/Qwen-14B}{\texttt{\textbf{Qwen/Qwen-14B}}}: Total 41 layers, 5120 neurons per layer. Sampled layers: 6, 13, 20, 27, 34.
    \item \href{https://huggingface.co/Qwen/Qwen-14B-Chat}{\texttt{\textbf{Qwen/Qwen-14B-Chat}}}: Total 41 layers, 5120 neurons per layer. Sampled layers: 6, 13, 20, 27, 34.
    \item \href{https://huggingface.co/Qwen/Qwen1.5-0.5B}{\texttt{\textbf{Qwen/Qwen1.5-0.5B}}}: Total 25 layers, 1024 neurons per layer. Sampled layers: 4, 8, 12, 16, 20.
    \item \href{https://huggingface.co/Qwen/Qwen1.5-0.5B-Chat}{\texttt{\textbf{Qwen/Qwen1.5-0.5B-Chat}}}: Total 25 layers, 1024 neurons per layer. Sampled layers: 4, 8, 12, 16, 20.
    \item \href{https://huggingface.co/Qwen/Qwen1.5-1.8B}{\texttt{\textbf{Qwen/Qwen1.5-1.8B}}}: Total 25 layers, 2048 neurons per layer. Sampled layers: 4, 8, 12, 16, 20.
    \item \href{https://huggingface.co/Qwen/Qwen1.5-1.8B-Chat}{\texttt{\textbf{Qwen/Qwen1.5-1.8B-Chat}}}: Total 25 layers, 2048 neurons per layer. Sampled layers: 4, 8, 12, 16, 20.
    \item \href{https://huggingface.co/Qwen/Qwen1.5-4B}{\texttt{\textbf{Qwen/Qwen1.5-4B}}}: Total 41 layers, 2560 neurons per layer. Sampled layers: 6, 13, 20, 27, 34.
    \item \href{https://huggingface.co/Qwen/Qwen1.5-4B-Chat}{\texttt{\textbf{Qwen/Qwen1.5-4B-Chat}}}: Total 41 layers, 2560 neurons per layer. Sampled layers: 6, 13, 20, 27, 34.
    \item \href{https://huggingface.co/Qwen/Qwen1.5-7B}{\texttt{\textbf{Qwen/Qwen1.5-7B}}}: Total 33 layers, 4096 neurons per layer. Sampled layers: 5, 11, 16, 22, 27.
    \item \href{https://huggingface.co/Qwen/Qwen1.5-7B-Chat}{\texttt{\textbf{Qwen/Qwen1.5-7B-Chat}}}: Total 33 layers, 4096 neurons per layer. Sampled layers: 5, 11, 16, 22, 27.
    \item \href{https://huggingface.co/Qwen/Qwen1.5-14B}{\texttt{\textbf{Qwen/Qwen1.5-14B}}}: Total 41 layers, 5120 neurons per layer. Sampled layers: 6, 13, 20, 27, 34.
    \item \href{https://huggingface.co/Qwen/Qwen1.5-14B-Chat}{\texttt{\textbf{Qwen/Qwen1.5-14B-Chat}}}: Total 41 layers, 5120 neurons per layer. Sampled layers: 6, 13, 20, 27, 34.
    \item \href{https://huggingface.co/Qwen/Qwen2-0.5B}{\texttt{\textbf{Qwen/Qwen2-0.5B}}}: Total 25 layers, 896 neurons per layer. Sampled layers: 4, 8, 12, 16, 20.
    \item \href{https://huggingface.co/Qwen/Qwen2-0.5B-Instruct}{\texttt{\textbf{Qwen/Qwen2-0.5B-Instruct}}}: Total 25 layers, 896 neurons per layer. Sampled layers: 4, 8, 12, 16, 20.
    \item \href{https://huggingface.co/Qwen/Qwen2-1.5B}{\texttt{\textbf{Qwen/Qwen2-1.5B}}}: Total 29 layers, 1536 neurons per layer. Sampled layers: 4, 9, 14, 19, 24.
    \item \href{https://huggingface.co/Qwen/Qwen2-1.5B-Instruct}{\texttt{\textbf{Qwen/Qwen2-1.5B-Instruct}}}: Total 29 layers, 1536 neurons per layer. Sampled layers: 4, 9, 14, 19, 24.
    \item \href{https://huggingface.co/Qwen/Qwen2-7B}{\texttt{\textbf{Qwen/Qwen2-7B}}}: Total 29 layers, 3584 neurons per layer. Sampled layers: 4, 9, 14, 19, 24.
    \item \href{https://huggingface.co/Qwen/Qwen2-7B-Instruct}{\texttt{\textbf{Qwen/Qwen2-7B-Instruct}}}: Total 29 layers, 3584 neurons per layer. Sampled layers: 4, 9, 14, 19, 24.
\end{itemize}

\textbf{SeaLLMs}
\begin{itemize}
    \item \href{https://huggingface.co/SeaLLMs/SeaLLMs-v3-1.5B}{\texttt{\textbf{SeaLLMs/SeaLLMs-v3-1.5B}}}: Total 29 layers, 1536 neurons per layer. Sampled layers: 4, 9, 14, 19, 24.
    \item \href{https://huggingface.co/SeaLLMs/SeaLLMs-v3-1.5B-Chat}{\texttt{\textbf{SeaLLMs/SeaLLMs-v3-1.5B-Chat}}}: Total 29 layers, 1536 neurons per layer. Sampled layers: 4, 9, 14, 19, 24.
    \item \href{https://huggingface.co/SeaLLMs/SeaLLMs-v3-7B}{\texttt{\textbf{SeaLLMs/SeaLLMs-v3-7B}}}: Total 29 layers, 3584 neurons per layer. Sampled layers: 4, 9, 14, 19, 24.
    \item \href{https://huggingface.co/SeaLLMs/SeaLLMs-v3-7B-Chat}{\texttt{\textbf{SeaLLMs/SeaLLMs-v3-7B-Chat}}}: Total 29 layers, 3584 neurons per layer. Sampled layers: 4, 9, 14, 19, 24.
\end{itemize}

\textbf{SmolLM}
\begin{itemize}
    \item \href{https://huggingface.co/HuggingFaceTB/SmolLM-135M}{\texttt{\textbf{HuggingFaceTB/SmolLM-135M}}}: Total 31 layers, 576 neurons per layer. Sampled layers: 5, 10, 15, 20, 25.
    \item \href{https://huggingface.co/HuggingFaceTB/SmolLM-135M-Instruct}{\texttt{\textbf{HuggingFaceTB/SmolLM-135M-Instruct}}}: Total 31 layers, 576 neurons per layer. Sampled layers: 5, 10, 15, 20, 25.
    \item \href{https://huggingface.co/HuggingFaceTB/SmolLM-360M}{\texttt{\textbf{HuggingFaceTB/SmolLM-360M}}}: Total 33 layers, 960 neurons per layer. Sampled layers: 5, 11, 16, 22, 27.
    \item \href{https://huggingface.co/HuggingFaceTB/SmolLM-360M-Instruct}{\texttt{\textbf{HuggingFaceTB/SmolLM-360M-Instruct}}}: Total 33 layers, 960 neurons per layer. Sampled layers: 5, 11, 16, 22, 27.
    \item \href{https://huggingface.co/HuggingFaceTB/SmolLM-1.7B}{\texttt{\textbf{HuggingFaceTB/SmolLM-1.7B}}}: Total 25 layers, 2048 neurons per layer. Sampled layers: 4, 8, 12, 16, 20.
    \item \href{https://huggingface.co/HuggingFaceTB/SmolLM-1.7B-Instruct}{\texttt{\textbf{HuggingFaceTB/SmolLM-1.7B-Instruct}}}: Total 25 layers, 2048 neurons per layer. Sampled layers: 4, 8, 12, 16, 20.
\end{itemize}

\textbf{TinyLlama}
\begin{itemize}
    \item \href{https://huggingface.co/TinyLlama/TinyLlama_v1.1}{\texttt{\textbf{TinyLlama/TinyLlama\_v1.1}}}: Total 23 layers, 2048 neurons per layer. Sampled layers: 3, 7, 11, 15, 19.
    \item \href{https://huggingface.co/TinyLlama/TinyLlama_v1.1_chinese}{\texttt{\textbf{TinyLlama/TinyLlama\_v1.1\_chinese}}}: Total 23 layers, 2048 neurons per layer. Sampled layers: 3, 7, 11, 15, 19.
    \item \href{https://huggingface.co/TinyLlama/TinyLlama_v1.1_math_code}{\texttt{\textbf{TinyLlama/TinyLlama\_v1.1\_math\_code}}}: Total 23 layers, 2048 neurons per layer. Sampled layers: 3, 7, 11, 15, 19.
\end{itemize}

\textbf{Yi}
\begin{itemize}
    \item \href{https://huggingface.co/01-ai/Yi-1.5-6B}{\texttt{\textbf{01-ai/Yi-1.5-6B}}}: Total 33 layers, 4096 neurons per layer. Sampled layers: 5, 11, 16, 22, 27.
    \item \href{https://huggingface.co/01-ai/Yi-1.5-6B-Chat}{\texttt{\textbf{01-ai/Yi-1.5-6B-Chat}}}: Total 33 layers, 4096 neurons per layer. Sampled layers: 5, 11, 16, 22, 27.
    \item \href{https://huggingface.co/01-ai/Yi-1.5-9B}{\texttt{\textbf{01-ai/Yi-1.5-9B}}}: Total 49 layers, 4096 neurons per layer. Sampled layers: 8, 16, 24, 32, 40.
    \item \href{https://huggingface.co/01-ai/Yi-1.5-9B-Chat}{\texttt{\textbf{01-ai/Yi-1.5-9B-Chat}}}: Total 49 layers, 4096 neurons per layer. Sampled layers: 8, 16, 24, 32, 40.
\end{itemize}

\textbf{Bloom \& Bloomz}
\begin{itemize}
    \item \href{https://huggingface.co/bigscience/bloom-560m}{\texttt{\textbf{bigscience/bloom-560m}}}: Total 25 layers, 1024 neurons per layer. Sampled layers: 4, 8, 12, 16, 20.
    \item \href{https://huggingface.co/bigscience/bloomz-560m}{\texttt{\textbf{bigscience/bloomz-560m}}}: Total 25 layers, 1024 neurons per layer. Sampled layers: 4, 8, 12, 16, 20.
    \item \href{https://huggingface.co/bigscience/bloom-1b1}{\texttt{\textbf{bigscience/bloom-1b1}}}: Total 25 layers, 1536 neurons per layer. Sampled layers: 4, 8, 12, 16, 20.
    \item \href{https://huggingface.co/bigscience/bloomz-1b1}{\texttt{\textbf{bigscience/bloomz-1b1}}}: Total 25 layers, 1536 neurons per layer. Sampled layers: 4, 8, 12, 16, 20.
    \item \href{https://huggingface.co/bigscience/bloom-1b7}{\texttt{\textbf{bigscience/bloom-1b7}}}: Total 25 layers, 2048 neurons per layer. Sampled layers: 4, 8, 12, 16, 20.
    \item \href{https://huggingface.co/bigscience/bloomz-1b7}{\texttt{\textbf{bigscience/bloomz-1b7}}}: Total 25 layers, 2048 neurons per layer. Sampled layers: 4, 8, 12, 16, 20.
    \item \href{https://huggingface.co/bigscience/bloom-3b}{\texttt{\textbf{bigscience/bloom-3b}}}: Total 31 layers, 2560 neurons per layer. Sampled layers: 5, 10, 15, 20, 25.
    \item \href{https://huggingface.co/bigscience/bloomz-3b}{\texttt{\textbf{bigscience/bloomz-3b}}}: Total 31 layers, 2560 neurons per layer. Sampled layers: 5, 10, 15, 20, 25.
    \item \href{https://huggingface.co/bigscience/bloom-7b1}{\texttt{\textbf{bigscience/bloom-7b1}}}: Total 31 layers, 4096 neurons per layer. Sampled layers: 5, 10, 15, 20, 25.
    \item \href{https://huggingface.co/bigscience/bloomz-7b1}{\texttt{\textbf{bigscience/bloomz-7b1}}}: Total 31 layers, 4096 neurons per layer. Sampled layers: 5, 10, 15, 20, 25.
\end{itemize}

\textbf{Gemma \& Gemma-2}
\begin{itemize}
    \item \href{https://huggingface.co/google/gemma-2b}{\texttt{\textbf{google/gemma-2b}}}: Total 19 layers, 2048 neurons per layer. Sampled layers: 3, 6, 9, 12, 15.
    \item \href{https://huggingface.co/google/gemma-2b-it}{\texttt{\textbf{google/gemma-2b-it}}}: Total 19 layers, 2048 neurons per layer. Sampled layers: 3, 6, 9, 12, 15.
    \item \href{https://huggingface.co/google/gemma-7b}{\texttt{\textbf{google/gemma-7b}}}: Total 29 layers, 3072 neurons per layer. Sampled layers: 4, 9, 14, 19, 24.
    \item \href{https://huggingface.co/google/gemma-7b-it}{\texttt{\textbf{google/gemma-7b-it}}}: Total 29 layers, 3072 neurons per layer. Sampled layers: 4, 9, 14, 19, 24.
    \item \href{https://huggingface.co/google/gemma-2-2b}{\texttt{\textbf{google/gemma-2-2b}}}: Total 27 layers, 2304 neurons per layer. Sampled layers: 4, 9, 13, 18, 22.
    \item \href{https://huggingface.co/google/gemma-2-2b-it}{\texttt{\textbf{google/gemma-2-2b-it}}}: Total 27 layers, 2304 neurons per layer. Sampled layers: 4, 9, 13, 18, 22.
    \item \href{https://huggingface.co/google/gemma-2-9b}{\texttt{\textbf{google/gemma-2-9b}}}: Total 43 layers, 3584 neurons per layer. Sampled layers: 7, 14, 21, 28, 35.
    \item \href{https://huggingface.co/google/gemma-2-9b-it}{\texttt{\textbf{google/gemma-2-9b-it}}}: Total 43 layers, 3584 neurons per layer. Sampled layers: 7, 14, 21, 28, 35.
\end{itemize}

\textbf{GLM}
\begin{itemize}
    \item \href{https://huggingface.co/THUDM/glm-4-9b}{\texttt{\textbf{THUDM/glm-4-9b}}}: Total 41 layers, 4096 neurons per layer. Sampled layers: 6, 13, 20, 27, 34.
    \item \href{https://huggingface.co/THUDM/glm-4-9b-chat}{\texttt{\textbf{THUDM/glm-4-9b-chat}}}: Total 41 layers, 4096 neurons per layer. Sampled layers: 6, 13, 20, 27, 34.
\end{itemize}

\textbf{GPT-Neo}
\begin{itemize}
    \item \href{https://huggingface.co/EleutherAI/gpt-neo-125m}{\texttt{\textbf{EleutherAI/gpt-neo-125m}}}: Total 13 layers, 768 neurons per layer. Sampled layers: 2, 4, 6, 8, 10.
    \item \href{https://huggingface.co/EleutherAI/gpt-neo-1.3B}{\texttt{\textbf{EleutherAI/gpt-neo-1.3B}}}: Total 25 layers, 2048 neurons per layer. Sampled layers: 4, 8, 12, 16, 20.
    \item \href{https://huggingface.co/EleutherAI/gpt-neo-2.7B}{\texttt{\textbf{EleutherAI/gpt-neo-2.7B}}}: Total 33 layers, 2560 neurons per layer. Sampled layers: 5, 11, 16, 22, 27.
    \item \href{https://huggingface.co/EleutherAI/gpt-neox-20B}{\texttt{\textbf{EleutherAI/gpt-neox-20B}}}: Total 45 layers, 6144 neurons per layer. Sampled layers: 7, 15, 22, 30, 37.
\end{itemize}

\textbf{GPT-2}
\begin{itemize}
    \item \href{https://huggingface.co/openai-community/gpt2}{\texttt{\textbf{openai-community/gpt2}}}: Total 13 layers, 768 neurons per layer. Sampled layers: 2, 4, 6, 8, 10.
    \item \href{https://huggingface.co/openai-community/gpt2-medium}{\texttt{\textbf{openai-community/gpt2-medium}}}: Total 25 layers, 1024 neurons per layer. Sampled layers: 4, 8, 12, 16, 20.
    \item \href{https://huggingface.co/openai-community/gpt2-large}{\texttt{\textbf{openai-community/gpt2-large}}}: Total 37 layers, 1280 neurons per layer. Sampled layers: 6, 12, 18, 24, 30.
    \item \href{https://huggingface.co/openai-community/gpt2-xl}{\texttt{\textbf{openai-community/gpt2-xl}}}: Total 49 layers, 1600 neurons per layer. Sampled layers: 8, 16, 24, 32, 40.
\end{itemize}

\textbf{InternLM-2.5}
\begin{itemize}
    \item \href{https://huggingface.co/internlm/internlm2_5-1_8b}{\texttt{\textbf{internlm/internlm2\_5-1\_8b}}}: Total 25 layers, 2048 neurons per layer. Sampled layers: 4, 8, 12, 16, 20.
    \item \href{https://huggingface.co/internlm/internlm2_5-1_8b-chat}{\texttt{\textbf{internlm/internlm2\_5-1\_8b-chat}}}: Total 25 layers, 2048 neurons per layer. Sampled layers: 4, 8, 12, 16, 20.
    \item \href{https://huggingface.co/internlm/internlm2_5-7b}{\texttt{\textbf{internlm/internlm2\_5-7b}}}: Total 33 layers, 4096 neurons per layer. Sampled layers: 5, 11, 16, 22, 27.
    \item \href{https://huggingface.co/internlm/internlm2_5-7b-chat}{\texttt{\textbf{internlm/internlm2\_5-7b-chat}}}: Total 33 layers, 4096 neurons per layer. Sampled layers: 5, 11, 16, 22, 27.
    \item \href{https://huggingface.co/internlm/internlm2_5-20b}{\texttt{\textbf{internlm/internlm2\_5-20b}}}: Total 49 layers, 6144 neurons per layer. Sampled layers: 8, 16, 24, 32, 40.
    \item \href{https://huggingface.co/internlm/internlm2_5-20b-chat}{\texttt{\textbf{internlm/internlm2\_5-20b-chat}}}: Total 49 layers, 6144 neurons per layer. Sampled layers: 8, 16, 24, 32, 40.
\end{itemize}

\textbf{OPT}
\begin{itemize}
    \item \href{https://huggingface.co/facebook/opt-125m}{\texttt{\textbf{facebook/opt-125m}}}: Total 13 layers, 768 neurons per layer. Sampled layers: 2, 4, 6, 8, 10.
    \item \href{https://huggingface.co/facebook/opt-1.3b}{\texttt{\textbf{facebook/opt-1.3b}}}: Total 25 layers, 2048 neurons per layer. Sampled layers: 4, 8, 12, 16, 20.
    \item \href{https://huggingface.co/facebook/opt-2.7b}{\texttt{\textbf{facebook/opt-2.7b}}}: Total 33 layers, 2560 neurons per layer. Sampled layers: 5, 11, 16, 22, 27.
    \item \href{https://huggingface.co/facebook/opt-6.7b}{\texttt{\textbf{facebook/opt-6.7b}}}: Total 33 layers, 4096 neurons per layer. Sampled layers: 5, 11, 16, 22, 27.
    \item \href{https://huggingface.co/facebook/opt-13b}{\texttt{\textbf{facebook/opt-13b}}}: Total 41 layers, 5120 neurons per layer. Sampled layers: 6, 13, 20, 27, 34.
\end{itemize}

\textbf{Phi}
\begin{itemize}
    \item \href{https://huggingface.co/microsoft/phi-1}{\texttt{\textbf{microsoft/phi-1}}}: Total 25 layers, 2048 neurons per layer. Sampled layers: 4, 8, 12, 16, 20.
    \item \href{https://huggingface.co/microsoft/phi-1_5}{\texttt{\textbf{microsoft/phi-1\_5}}}: Total 25 layers, 2048 neurons per layer. Sampled layers: 4, 8, 12, 16, 20.
    \item \href{https://huggingface.co/microsoft/phi-2}{\texttt{\textbf{microsoft/phi-2}}}: Total 33 layers, 2560 neurons per layer. Sampled layers: 5, 11, 16, 22, 27.
    \item \href{https://huggingface.co/microsoft/Phi-3-mini-128k-instruct}{\texttt{\textbf{microsoft/Phi-3-mini-128k-instruct}}}: Total 33 layers, 3072 neurons per layer. Sampled layers: 5, 11, 16, 22, 27.
    \item \href{https://huggingface.co/microsoft/Phi-3-small-128k-instruct}{\texttt{\textbf{microsoft/Phi-3-small-128k-instruct}}}: Total 33 layers, 4096 neurons per layer. Sampled layers: 5, 11, 16, 22, 27.
\end{itemize}

\textbf{Pythia}
\begin{itemize}
    \item \href{https://huggingface.co/EleutherAI/pythia-1.4b}{\texttt{\textbf{EleutherAI/pythia-1.4b}}}: Total 25 layers, 2048 neurons per layer. Sampled layers: 4, 8, 12, 16, 20.
    \item \href{https://huggingface.co/EleutherAI/pythia-12b}{\texttt{\textbf{EleutherAI/pythia-12b}}}: Total 37 layers, 5120 neurons per layer. Sampled layers: 6, 12, 18, 24, 30.
    \item \href{https://huggingface.co/EleutherAI/pythia-14m}{\texttt{\textbf{EleutherAI/pythia-14m}}}: Total 7 layers, 128 neurons per layer. Sampled layers: 1, 2, 3, 4, 5.
    \item \href{https://huggingface.co/EleutherAI/pythia-160m}{\texttt{\textbf{EleutherAI/pythia-160m}}}: Total 13 layers, 768 neurons per layer. Sampled layers: 2, 4, 6, 8, 10.
    \item \href{https://huggingface.co/EleutherAI/pythia-1b}{\texttt{\textbf{EleutherAI/pythia-1b}}}: Total 17 layers, 2048 neurons per layer. Sampled layers: 2, 5, 8, 11, 14.
    \item \href{https://huggingface.co/EleutherAI/pythia-2.8b}{\texttt{\textbf{EleutherAI/pythia-2.8b}}}: Total 33 layers, 2560 neurons per layer. Sampled layers: 5, 11, 16, 22, 27.
    \item \href{https://huggingface.co/EleutherAI/pythia-410m}{\texttt{\textbf{EleutherAI/pythia-410m}}}: Total 25 layers, 1024 neurons per layer. Sampled layers: 4, 8, 12, 16, 20.
    \item \href{https://huggingface.co/EleutherAI/pythia-6.7b}{\texttt{\textbf{EleutherAI/pythia-6.7b}}}: Total 33 layers, 4096 neurons per layer. Sampled layers: 5, 11, 16, 22, 27.
    \item \href{https://huggingface.co/EleutherAI/pythia-70m}{\texttt{\textbf{EleutherAI/pythia-70m}}}: Total 7 layers, 512 neurons per layer. Sampled layers: 1, 2, 3, 4, 5.
\end{itemize}

\textbf{Others}
\begin{itemize}
    \item \href{https://huggingface.co/anas-awadalla/mpt-1b-redpajama-200b}{\texttt{\textbf{anas-awadalla/mpt-1b-redpajama-200b}}}: Total 24 layers, 2048 neurons per layer. Sampled layers: 4, 8, 12, 16, 20.
    \item \href{https://huggingface.co/CohereForAI/aya-23-8B}{\texttt{\textbf{CohereForAI/aya-23-8B}}}: Total 33 layers, 4096 neurons per layer. Sampled layers: 5, 11, 16, 22, 27.
    \item \href{https://huggingface.co/distilbert/distilgpt2}{\texttt{\textbf{distilbert/distilgpt2}}}: Total 7 layers, 768 neurons per layer. Sampled layers: 1, 2, 3, 4, 5.
\end{itemize}
}

\section{List of Linguistic Phenomena used in Section~\ref{sec:across-languages}}
\label{appendix:full-list-of-phenomena}

{\small
\textbf{BLIMP Dataset (English)}
\begin{itemize}

\item Anaphor agreement

    \begin{itemize}
    \item \texttt{anaphor gender agreement}
    \item \texttt{anaphor number agreement}
    \end{itemize}

\item Polarity item
    \begin{itemize}
    \item \texttt{matrix question npi licensor present}
    \item \texttt{npi present 1}
    \item \texttt{npi present 2}
    \item \texttt{only npi licensor present}
    \item \texttt{only npi scope}
    \item \texttt{sentential negation npi licensor present}
    \item \texttt{sentential negation npi scope}
    \end{itemize}

\item Filler gap dependency
    \begin{itemize}
    \item \texttt{wh questions object gap}
    \item \texttt{wh questions subject gap}
    \item \texttt{wh questions subject gap long distance}
    \item \texttt{wh vs that no gap}
    \item \texttt{wh vs that no gap long distance}
    \item \texttt{wh vs that with gap}
    \item \texttt{wh vs that with gap long distance}
    \end{itemize}
\end{itemize}

\textbf{SLING Dataset (Chinese)}
\begin{itemize}

    \item Anaphor agreement
    \begin{itemize}
    \item \texttt{self male}
    \item \texttt{cl men self female}
    \item \texttt{pp male}
    \item \texttt{baseline male}
    \item \texttt{baseline cl men female}
    \item \texttt{baseline cl female}
    \item \texttt{cl self male}
    \item \texttt{cl men self male}
    \item \texttt{self female}
    \item \texttt{pp female}
    \item \texttt{baseline cl men male}
    \item \texttt{menself female}
    \item \texttt{menself male}
    \item \texttt{cl self female}
    \item \texttt{baseline cl male}
    \item \texttt{baseline female}
    \item \texttt{baseline men female}
    \item \texttt{baseline men male}
    \end{itemize}

    \item Polarity item
    \begin{itemize}
    \item \texttt{any}
    \item \texttt{more or less}
    \item \texttt{even wh}
    \end{itemize}

    \item Filler gap dependency
    \begin{itemize}
    \item \texttt{rc resumptive noun}
    \item \texttt{rc resumptive pronoun}
    \end{itemize}
\end{itemize}
}

\onecolumn
\newpage
\section{Additional Visualizations}

\subsection{Consistency of Linguistic Similarity Across 104 LLMs}
\label{appendix:llm_alignment}

\begin{figure*}[h!]
    \centering
    \includegraphics[width=1\linewidth]{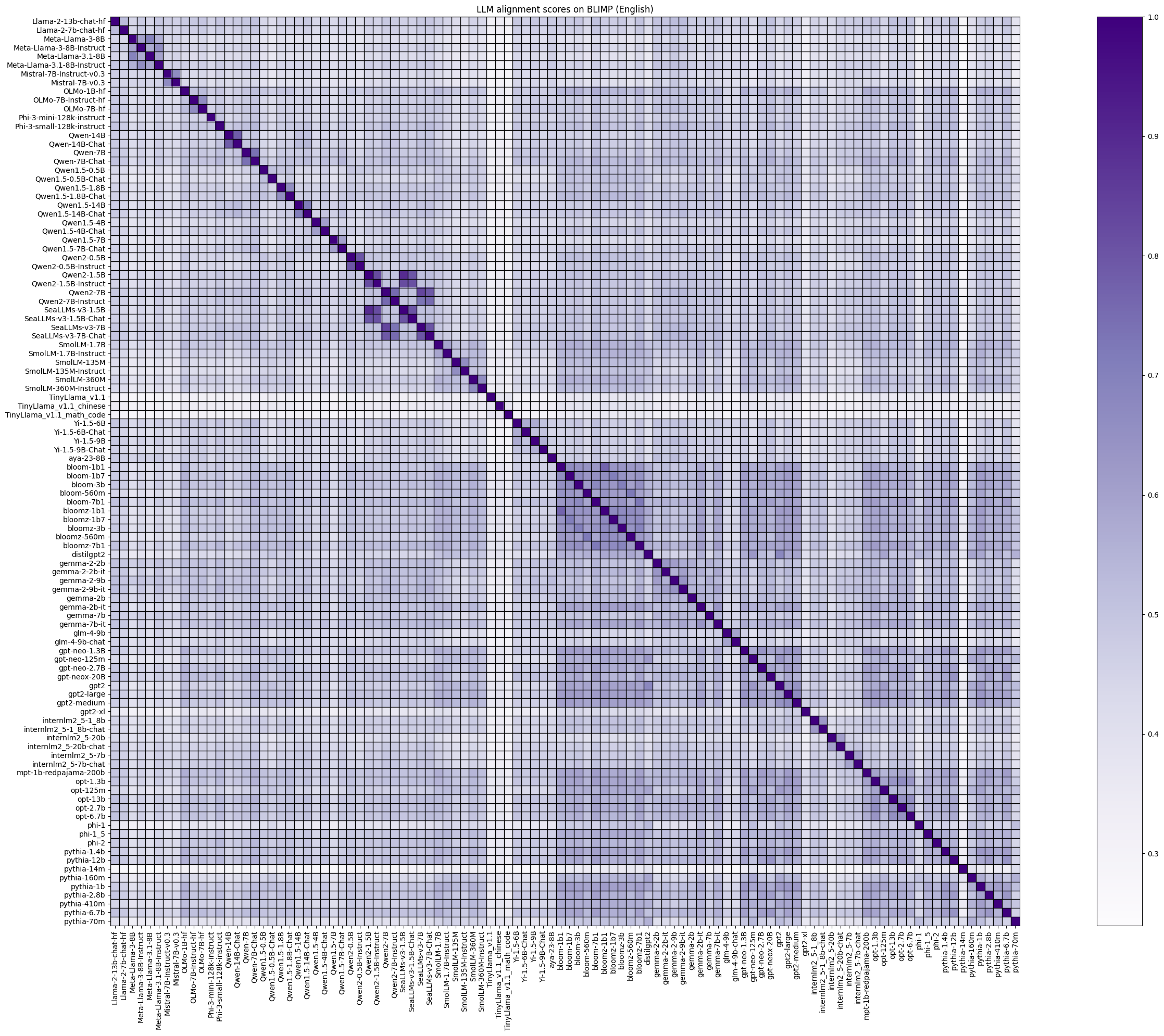}
    \caption{Pair-wise alignment scores of 104 LLMs in BLiMP dataset (English), corresponding to Fig.~\ref{fig:LLM_alignments} left.}
\end{figure*}

\begin{figure*}
    \centering
    \includegraphics[width=1\linewidth]{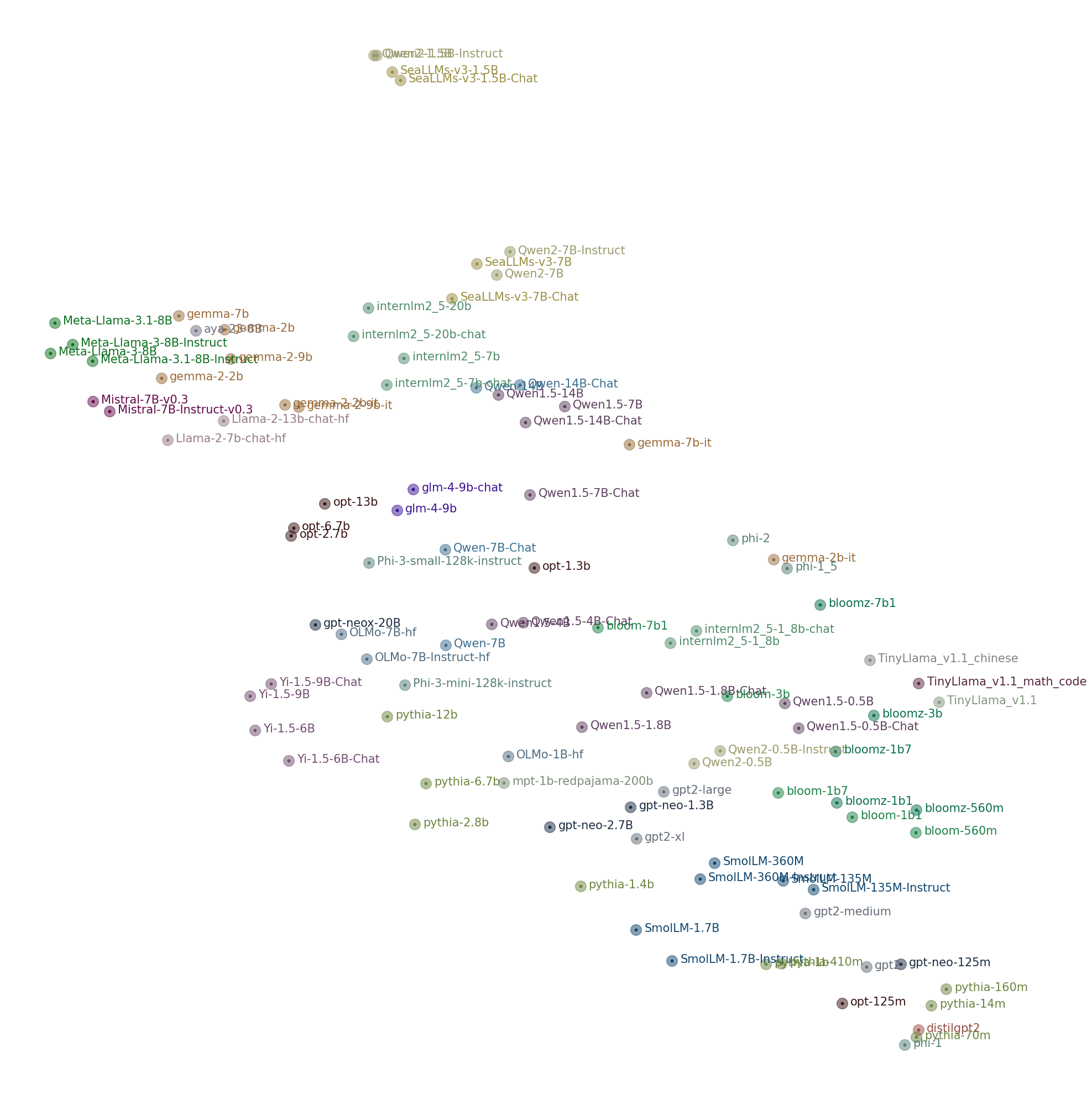}
    \caption{UMAP visualization based on LLM alignment scores on BLiMP dataset (English), corresponding to Fig.~\ref{fig:LLM_alignments} left.}
\end{figure*}

\begin{figure*}
    \centering
    \includegraphics[width=1\linewidth]{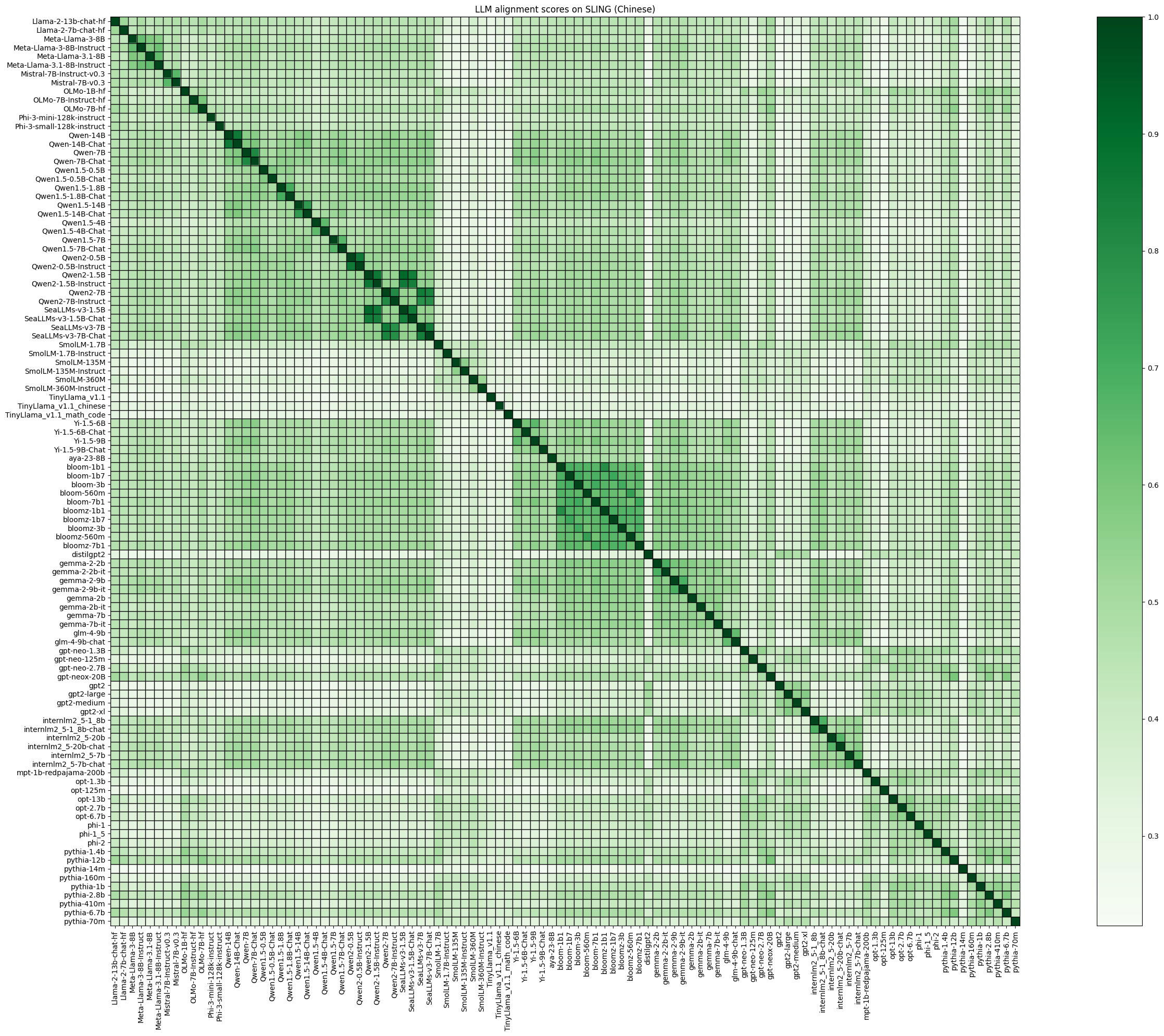}
    \caption{Pair-wise alignment scores of 104 LLMs in SLING dataset (Chinese), corresponding to Fig.~\ref{fig:LLM_alignments} right.}
\end{figure*}

\begin{figure*}
    \centering
    \includegraphics[width=1\linewidth]{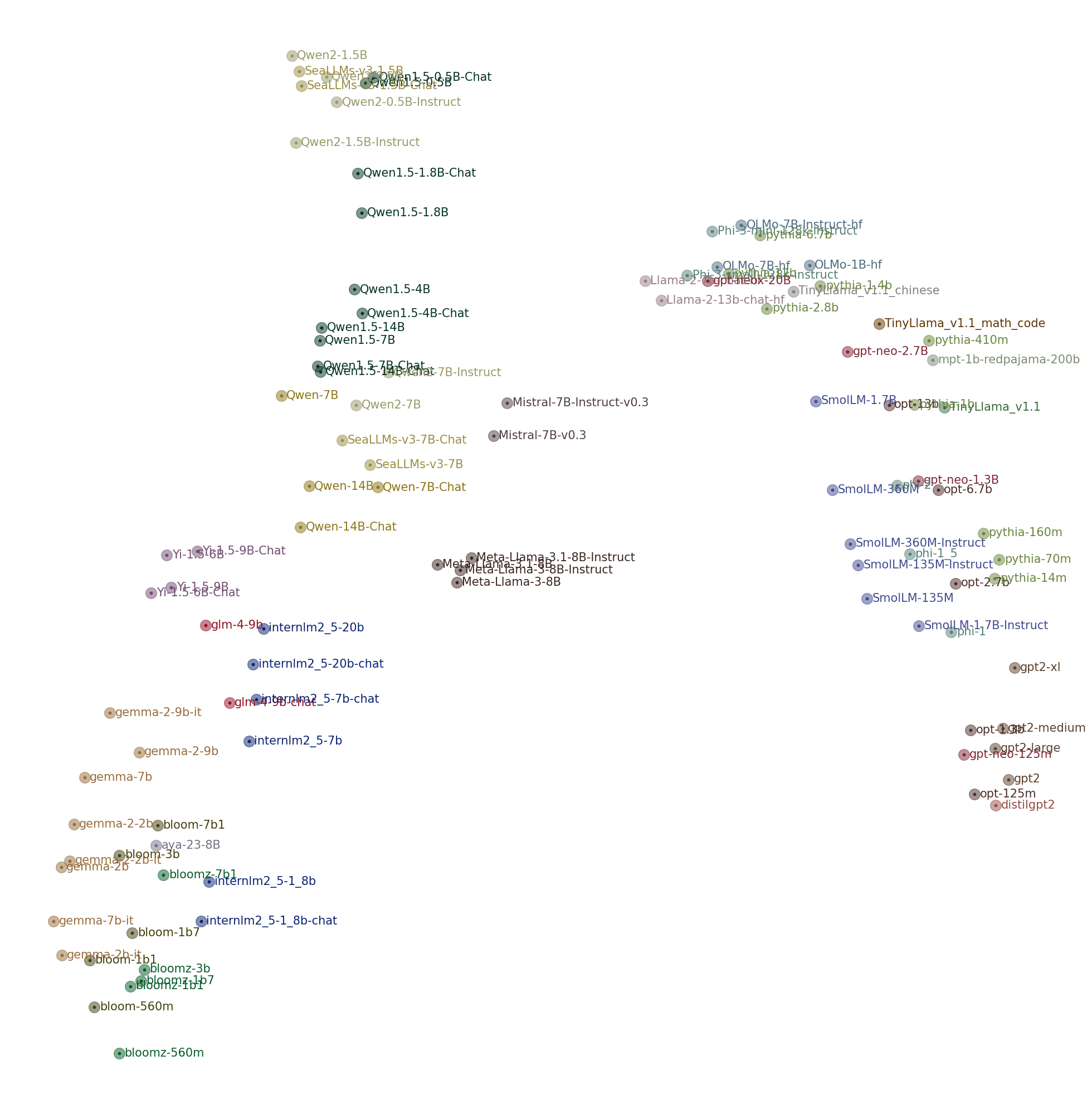}
    \caption{UMAP visualization based on LLM alignment scores on SLING dataset (Chinese), corresponding to Fig.~\ref{fig:LLM_alignments} right.}
\end{figure*}

\begin{figure*}
    \centering
    \includegraphics[width=1\linewidth]{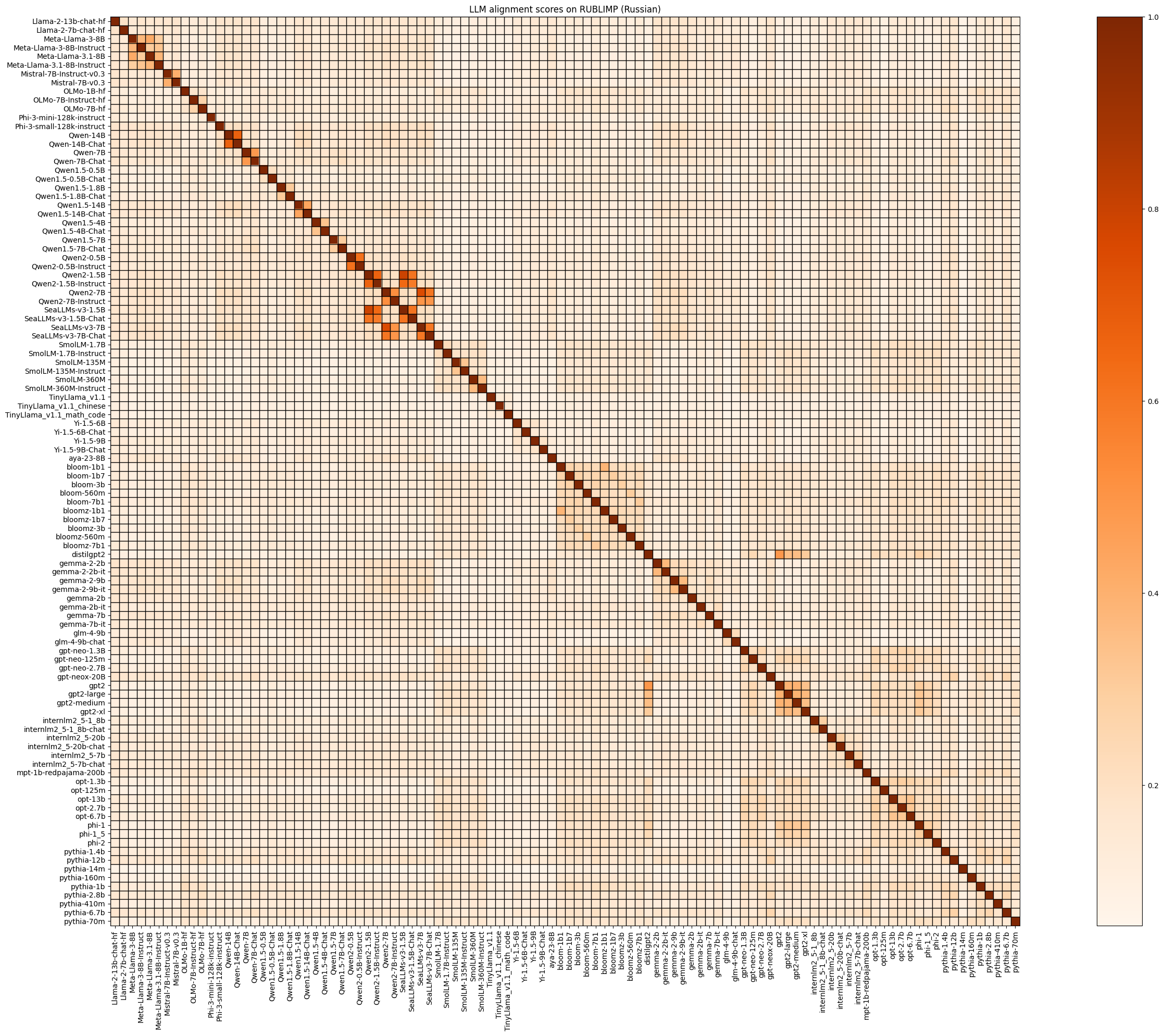}
    \caption{Pair-wise alignment scores of 104 LLMs in RuBLiMP dataset (Russian).}
\end{figure*}

\begin{figure*}
    \centering
    \includegraphics[width=1\linewidth]{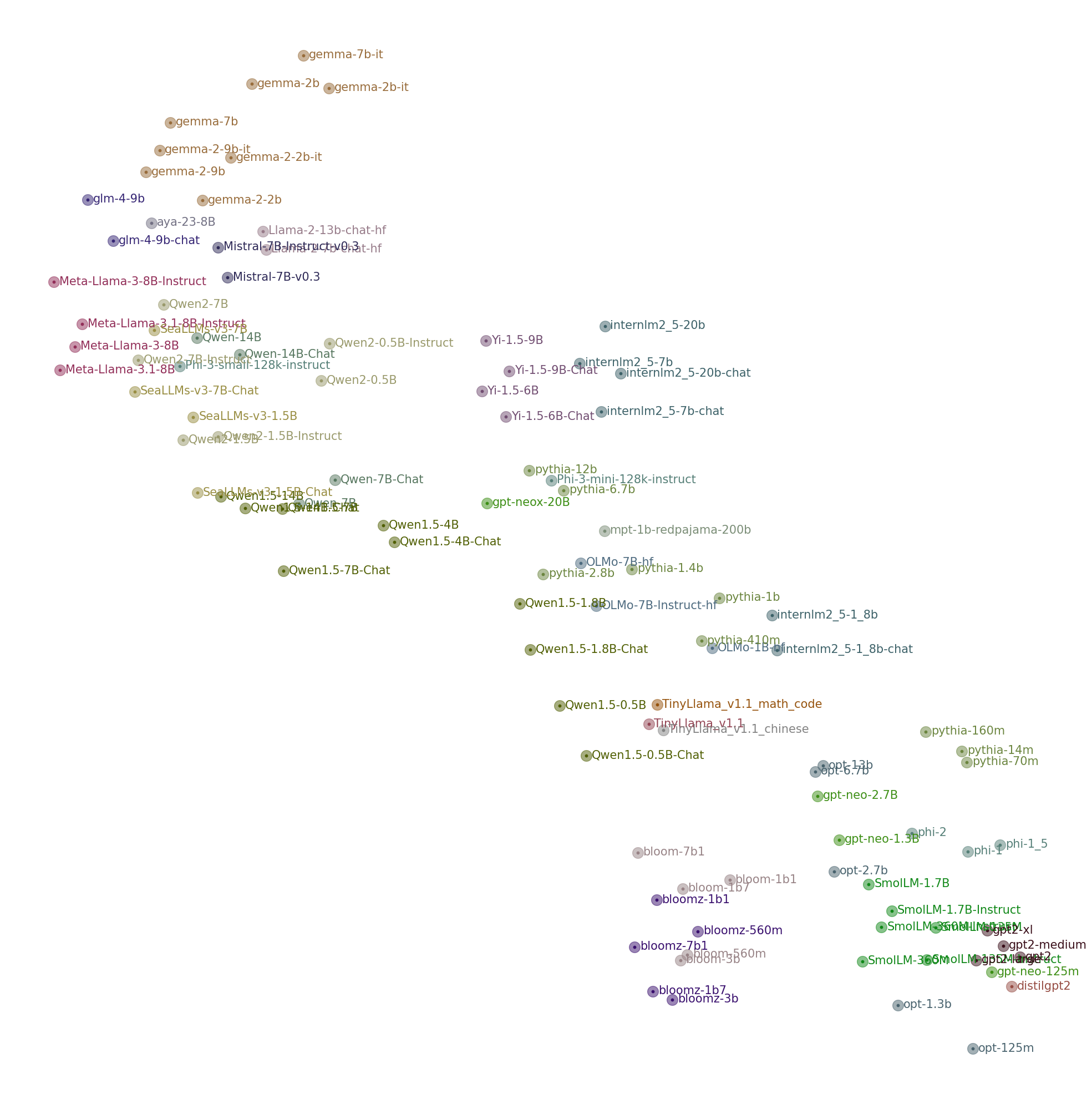}
    \caption{UMAP visualization based on LLM alignment scores on RuBLiMP dataset (Russian).}
\end{figure*}

\clearpage
\newpage
\subsection{Alignment between Linguistic Similarity and Theoretical Categorizations}
\label{appendix:Phenomena Similarities}

\begin{figure*}[h!]
    \centering
    \includegraphics[width=1\linewidth]{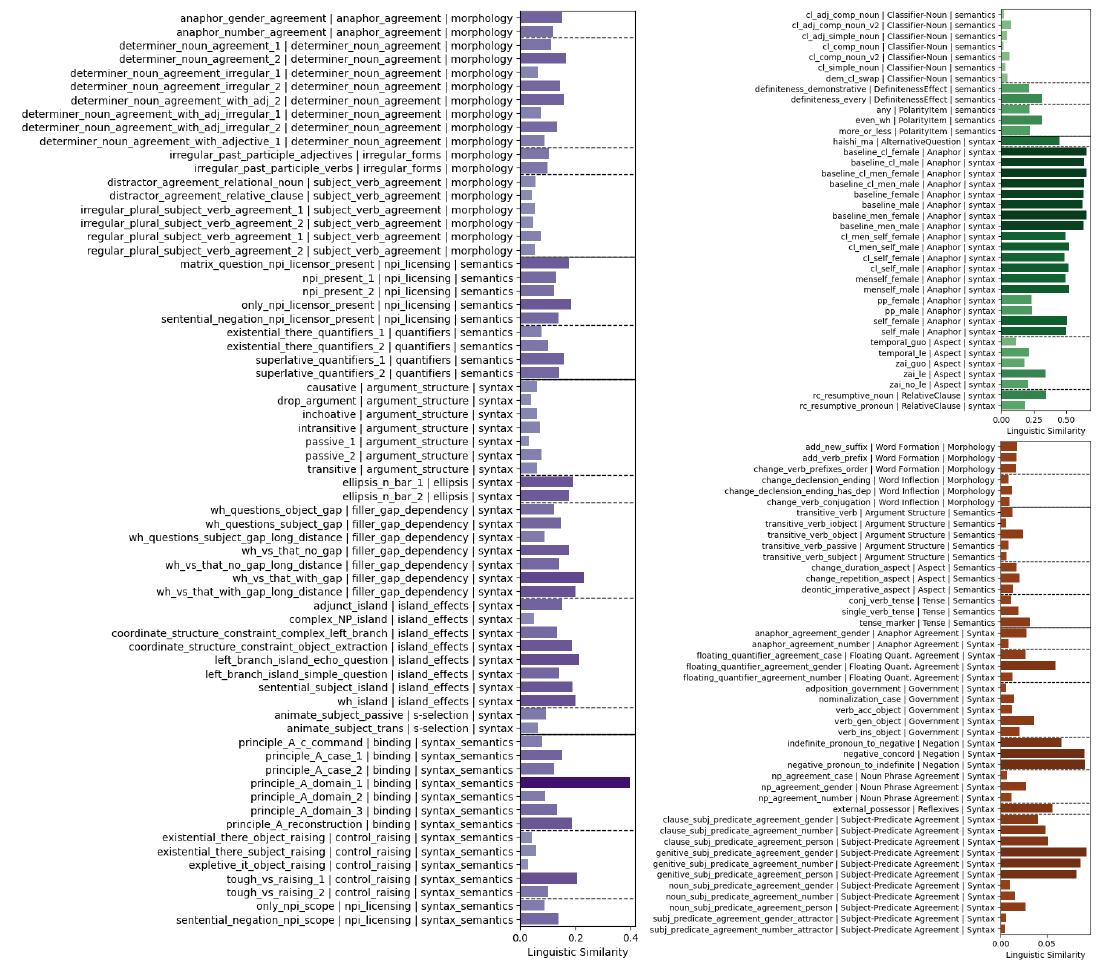}
    \caption{Self-similarities of linguistic phenomena in English, Chinese, and Russian}
\end{figure*}

\begin{figure*}
    \centering
    \includegraphics[width=1\linewidth]{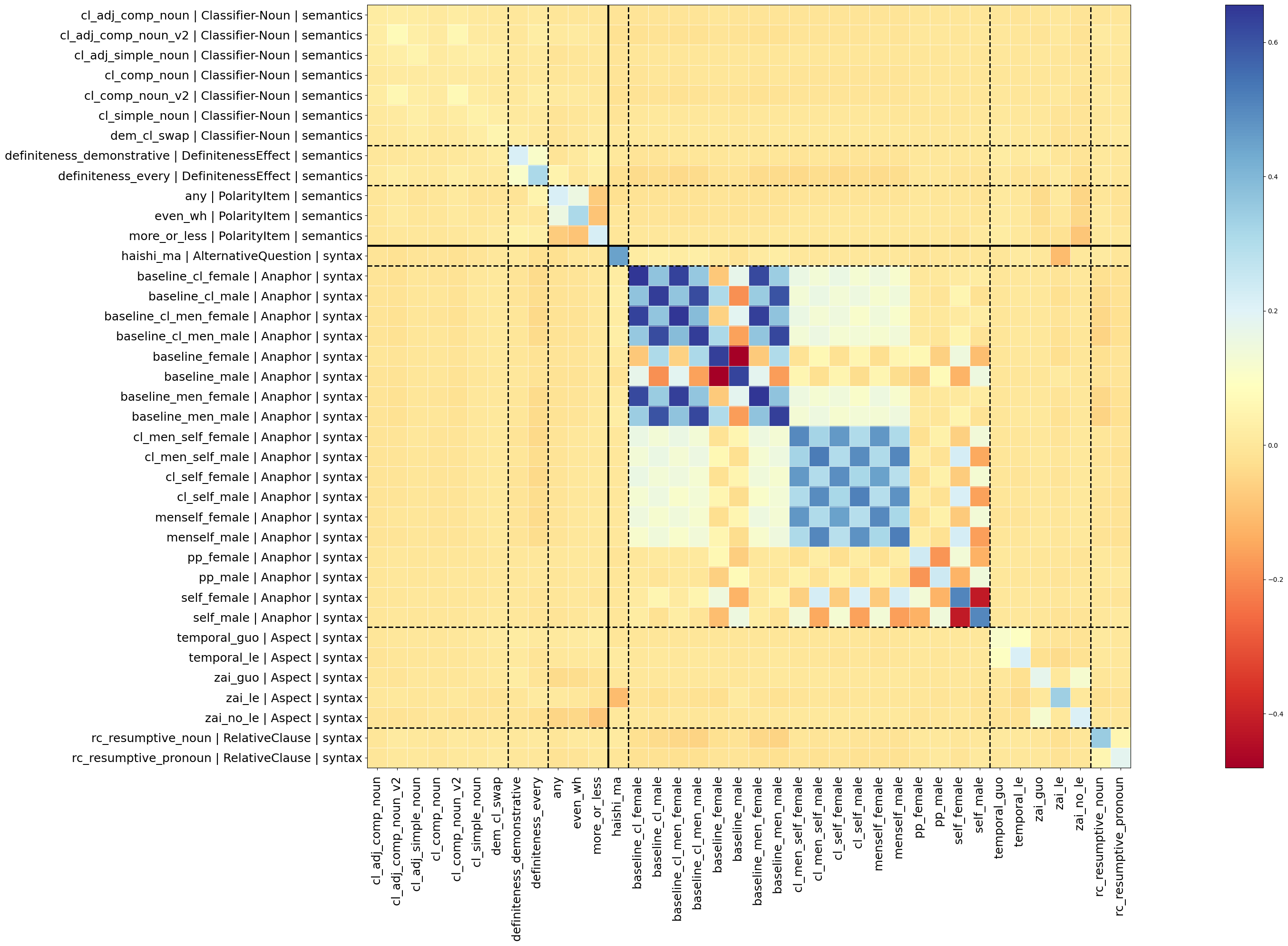}
    \caption{Phenomena-level linguistic similarity matrix of SLING}
\end{figure*}

\begin{figure*}
    \centering
    \includegraphics[width=1\linewidth]{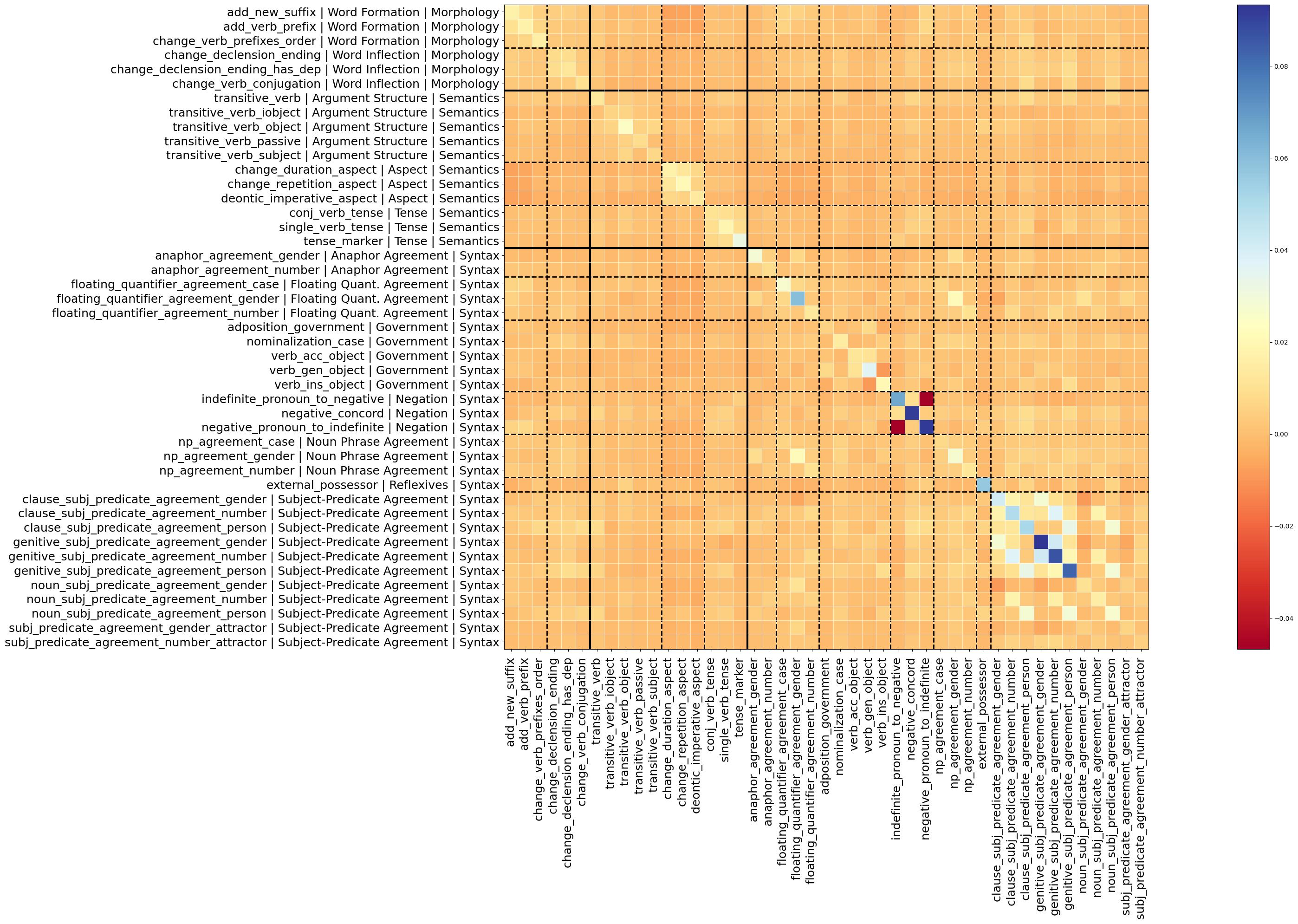}
    \caption{Phenomena-level linguistic similarity matrix of RUBLiMP}
\end{figure*}

\clearpage
\newpage
\subsection{Linguistic Similarity Across Different Languages}
\label{appendix:cross-lingual}

\begin{figure*}[h!]
    \centering
    \includegraphics[width=\linewidth]{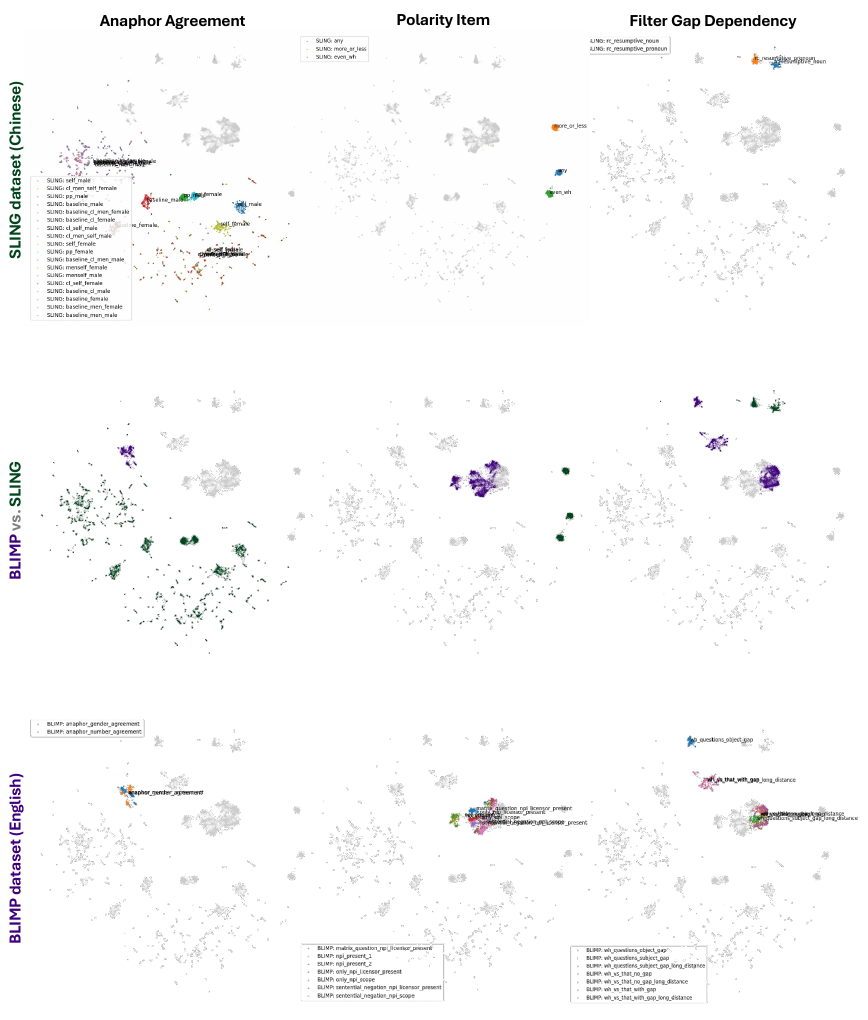}
    \caption{Extended detailed visualization of Fig.~\ref{fig:cross_lingual_umap}.}
\end{figure*}

\end{document}